\documentclass[sigconf]{acmart}
\usepackage{graphicx}
\usepackage{xcolor}
\usepackage{multirow}
\usepackage{tabularx}
\usepackage{algorithm}
\usepackage{algorithmic}
\usepackage{subcaption}
\usepackage{amsthm}
\usepackage{enumitem}

\newcolumntype{C}{>{\centering\arraybackslash}X} 

\copyrightyear{2022}
\acmYear{2022}
\setcopyright{acmcopyright}\acmConference[WSDM '22]{Proceedings of the Fifteenth
ACM International Conference on Web Search and Data Mining}{February 21--25,
2022}{Tempe, AZ, USA}
\acmBooktitle{Proceedings of the Fifteenth ACM International Conference on Web
Search and Data Mining (WSDM '22), February 21--25, 2022, Tempe, AZ, USA}
\acmPrice{15.00}
\acmDOI{10.1145/3488560.3498408}
\acmISBN{978-1-4503-9132-0/22/02}

\title{Towards Robust Graph Neural Networks for Noisy Graphs with Sparse Labels}

\author{Enyan Dai$^\dagger$, Wei Jin$^\ddagger$, Hui Liu$^\ddagger$, Suhang Wang$^\dagger$ }

\affiliation{$\dagger$ The Pennsylvania State University,
{${\ddagger}$} Michigan State University
}

\email{{emd5759, szw494}@psu.edu, {jinwei2, liuhui7}@msu.edu}
\begin{document}
\fancyhead{}
\begin{abstract}
Graph Neural Networks (GNNs) have shown their great ability in modeling graph structured data. However, real-world graphs usually contain structure noises and have limited labeled nodes. The performance of GNNs would drop significantly when trained on such graphs, which hinders the adoption of GNNs on many applications. Thus, it is important to develop noise-resistant GNNs with limited labeled nodes. However, the work on this is rather limited. Therefore, we study a novel problem of developing robust GNNs on noisy graphs with limited labeled nodes. Our analysis shows that both the noisy edges and limited labeled nodes could harm the message-passing mechanism of GNNs. To mitigate these issues, we propose a novel framework which adopts the noisy edges as supervision to learn a denoised and dense graph, which can down-weight or eliminate noisy edges and facilitate message passing of GNNs to alleviate the issue of limited labeled nodes. The generated edges are further used to 
regularize the predictions of unlabeled nodes with label smoothness to better train GNNs. Experimental results on real-world datasets demonstrate the robustness of the proposed framework on noisy graphs with limited labeled nodes.

\end{abstract}


\maketitle

\section{Introduction}
Graph Neural Networks (GNNs)~\cite{kipf2016semi,hamilton2017inductive} have made remarkable achievements in modeling graphs from various domains such as social networks~\cite{hamilton2017inductive}, financial system~\cite{wang2019semi}, and recommendation system~\cite{wang2019knowledge}. The success of GNNs relies on the message-passing mechanism~\cite{kipf2016semi,hamilton2017inductive}, where node representations are updated by aggregating the information from neighbors. With this mechanism, the node representations capture node features, information of neighbors and local graph structure, which facilitate various tasks, especially semi-supervised node classification. 

\begin{figure}
    \centering
    \includegraphics[width=0.95\linewidth]{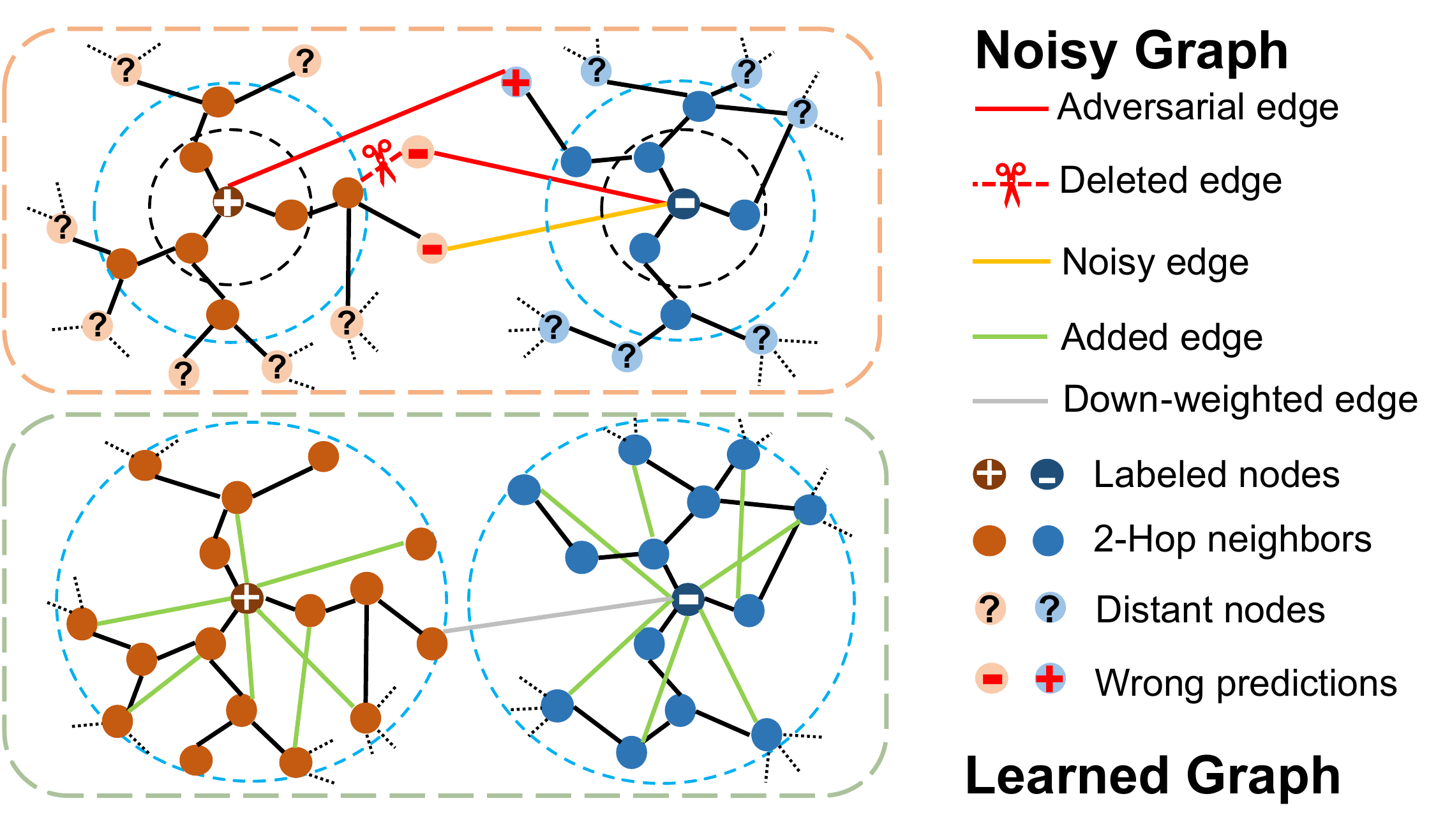}
    \vskip -1.9em
    \caption{An illustration of down-weighting/removing noise edges and densifying the graph for better performance.}
    \label{fig:illustration}
    \vskip -2em
\end{figure}

Although GNNs have shown great ability in modeling graphs, their performance can degrade significantly when trained on graphs with \textit{noisy edges} and/or \textit{limited labeled nodes}.
\textit{First}, due to the message passing, GNNs are vulnerable to adversarial or noisy edges. For example, as shown in Fig.~\ref{fig:illustration}, poisoning attacks~\cite{zugner2019adversarial} add/delete carefully chosen edges to the graph. These adversarial edges (shown in red) usually connect nodes of different labels or features, thus contaminating the neighborhoods of nodes, propagating noises/errors to node representations. In addition, inherent edge noises also exist in real-world graphs. For instance, in social networks, bots tend to build connections with normal users to spread misinformation~\cite{ferrara2016rise}, which can also harm the performance of GNNs for bot detection. 
\textit{Second}, for many applications, graphs are often sparsely labeled such as cell phone network for fraud detection~\cite{gallagher2008using}. 
Label sparsity can severely reduce the involvement of unlabeled nodes during message passing, leading to poor performance. Generally, in a $K$-layer GNN, a labeled node aggregates its $K$-hop neighborhood information, thus making many unlabeled nodes in $K$-hop neighborhood participate in the training, which is one major reason that GNNs can leverage unlabeled nodes for semi-supervised node classification. However,  as verified in our preliminary analysis in Fig.~\ref{fig:1_a} of Sec.~\ref{sec:3_3}, when the number of labeled nodes decreases, the amount of unlabeled nodes participating in training drops quickly, making message passing less effective.
These shortcomings of GNNs hinder the adoption of GNNs for many real-world applications. Thus, it is important to develop robust GNNs that can simultaneously handle noisy graphs with sparse labels.

However, developing robust GNNs for graphs with noisy edges and limited labeled nodes is challenging. \textit{First}, the training graph itself is noisy, i.e., noisy edges are mixed with the normal edges. Thus, we need supervision in down-weighting or eliminating noisy edges. \textit{Second}, alleviating the limited label issue requires more labels, while obtaining more labeled nodes is time-consuming and expensive. Hence, we need alternative approaches to more effectively utilize the limited labels. Some initial efforts~\cite{wu2019adversarial,jin2020graph,tang2020transferring,jin2020graph} have been taken to alleviate the effects of the adversarial edges such as pruning edges by using node similarity~\cite{wu2019adversarial}, and adopting Gaussian distribution as node representations to absorb noises~\cite{zhu2019robust}. To address the problem of sparsely labeled graphs, some methods~\cite{sun2019multi,li2018deeper,peng2020self} propose to obtain better representations by training GNNs with self-supervised learning tasks such as pseudo label prediction~\cite{sun2019multi,li2018deeper} and global context predictions~\cite{peng2020self}. 
However, little efforts are taken for robust GNNs that can simultaneously handle noisy edges and label sparsity.

Since both the noisy edges and limited labeled nodes harm the message passing of GNNs and message passing is directly related to the graph structure, we argue that learning a denoised and dense graph guided by the raw attributed graph is promising to facilitate message passing for robust GNNs. \textit{First}, for many graphs such as social networks, nodes with similar features and labels tend to be linked~\cite{liben2007link}, while noisy edges would link nodes of dissimilar features~\cite{wu2019adversarial}. 
Thus, we can use node attributes to predict the links. For existing links, the link predictor will assign small weights to links connecting nodes of dissimilar features while large weights to links connecting nodes of similar features, thus alleviating negative issue of noisy edges during message passing. \textit{Second}, 
real-world graphs are usually very sparse, containing many missing edges. With the link predictor, nodes that are potentially to be linked could be identified. Densifying the graph by linking similar nodes would induce more unlabeled nodes to become neighbors of labeled nodes with the same labels as shown in Fig.~\ref{fig:illustration}, which can alleviate the label sparsity issue. 
In addition, since adjacent nodes tend to have the same labels, the predicted new links can be used to further regularize the label predictions of unlabeled nodes. Though promising, the work on down-weighting noisy edges and densifying graph for robust GNN on noisy graphs with sparse labels are rather limited. 

Therefore, in this paper, we investigate a novel problem of developing robust noise-resistant GNNs with limited labeled nodes by learning a denoised and densified graph. In essence, we need to solve two challenges: (i) how to effectively learn a link predictor from the noisy graph which can eliminate noisy edges and densify the graph; and (ii) how to simultaneously use the learned graph to learn a structural noise-resistant GNNs with limited labeled nodes. To address these challenges, we propose a novel framework named 
robust structural noise-resistant GNN (RS-GNN)~\footnote{Codes are available at: https://github.com/EnyanDai/RSGNN}. RS-GNN adopts the node attributes and supervision from the noisy edges to
denoise and dense graph, which can alleviate the negative effects of noisy edges and facilitate the message passing between unlabeled nodes and labeled nodes. The learned graph is used as input for learning a GNN. RS-GNN also adopts the predicted edges to further explicitly regularize the predictions of unlabeled nodes to alleviate the label sparsity issue. In summary, our main contributions are:
\begin{itemize}[leftmargin=*]
    \item We study a new problem of learning robust noise-resistant GNNs with limited labeled nodes;
    \item We propose a novel framework RS-GNN, which can simultaneously learn a denoised and densified graph and a robust GNN on noisy graphs with limited labeled nodes; and
    \item We conduct extensive experiments on real-world datasets to demonstrate the robustness of RS-GNN on both noisy/clean graphs with limited labeled nodes. 
\end{itemize}

\section{Related Work}
\label{sec:related_work}

\subsection{Graph Neural Networks}
Graph Neural Networks (GNNs) have shown their great power in modeling graph structured data for various applications~\cite{wang2019semi,wang2018cross,zhao2020semi,dai2021say,zhao2021graphsmote}.
To generalize neural networks for graphs, two categories of GNNs are proposed, i.e., spectral-based~\cite{bruna2013spectral,henaff2015deep,kipf2016semi,levie2018cayleynets} and spatial-based~\cite{velivckovic2017graph,hamilton2017inductive,chen2018fastgcn,chiang2019cluster}. \citeauthor{bruna2013spectral} \cite{bruna2013spectral} first propose spectral-based GNNs by defining graph convolution with spectral graph theory. For instance, GCN~\cite{kipf2016semi} simplifies the convolutional operation by using the first order approximation. Spatial-based graph convolution is defined in spatial domain, which updates node representation by aggregating its neighbors' representations \cite{niepert2016learning,gilmer2017neural,hamilton2017inductive}. 
For example, self-attention of neighbor nodes is leveraged in graph attention network (GAT) \cite{velivckovic2017graph}. Moreover, various spatial methods are proposed to solve the scalability issue~\cite{chen2018fastgcn,chiang2019cluster} and learn deeper GNNs~\cite{chen2020simple}.  Recently, to alleviate the problem of lacking labeled nodes, many efforts are taken to explore GNNs using self-supervision, which aims to learn better node representations with pretext tasks~\cite{sun2019multi,li2018deeper,kim2021find,zhu2020self,jin2020self,dai2021towards}. For instance, superGAT~\cite{kim2021find} deploys edge prediction in GAT to guide the learning of attention for better representations. SE-GNN~\cite{dai2021towards} deploys contrastive learning to benefit the similarity modeling for self-explainable GNN.

Inspired by the great success of GNNs, methods that construct graphs and adopt GNNs for data without explicit relational structure are also explored~\cite{henaff2015deep,chen2019multi,jiang2019semi,dai2021nrgnn}. Generally, a graph would be built based on certain rules~\cite{henaff2015deep,chen2019multi} or be learned in an end-to-end model~\cite{jiang2019semi,dai2021nrgnn}. Our RS-GNN is inherently different from these methods as we eliminate/down-weight the noisy edges and predict the missing edges for robust GNNs on noisy graphs with limited labels. 

\subsection{Robust GNNs}
Although GNNs have obtained great achievements, they are vulnerable to adversarial attacks~\cite{wu2019adversarial,dai2018adversarial,zugner2018adversarial,zugner2019adversarial}. Based on the objective, the adversarial attacks on GNNs can be split into two categories, i.e., targeted attack~\cite{dai2018adversarial,zugner2018adversarial} and non-targeted attack~\cite{zugner2019adversarial}. Targeted attack methods aim to degrade the performance of the GNNs on target nodes. 
For instance, \textit{nettack}~\cite{zugner2018adversarial} adds adversarial perturbations to a graph to attack targeted nodes. Non-targeted attack aims to reduce the overall performance of GNNs. For example, \textit{metattack}~\cite{zugner2019adversarial} poisons the graph globally to achieve non-targeted attack with meta-learning. To defend against adversarial attacks, many efforts are taken recently~\cite{zhu2019robust,wu2019adversarial,entezari2020all,jin2020graph,tang2020transferring,zhang2020gnnguard}. \cite{wu2019adversarial} prune the perturbed edges based on Jaccard similarity of node features. Another preprocessing method by low-rank approximation of adjacent matrix is investigated~\cite{entezari2020all}. Pro-GNN~\cite{jin2020graph} is the most similar work to ours, which learns a clean graph structure by low-rank constraint. However, they only tackle the adversarial edges and their computational cost is very large due to the direct learning of the graph and the sparse low-rank constraint.
This work is inherently different from these methods as: (i) we study a novel problem of developing robust GNN for both noisy graphs and label sparsity issues; and (ii) the proposed RS-GNN simultaneously tackles the two issues by learning an link predictor to 
down-weight noisy edges and connecting nodes with high similarity to facilitate message-passing; 
and (iii) RS-GNN uses link predictor instead of direct graph learning to save computational cost. 
\section{Preliminary Analysis}
\label{Sec:pre_analysis}
In this section, we discuss the inner working of GNNs, conduct preliminary analysis to show the issues of GNN with sparse labels and verify that densifying graphs by connecting similar nodes can potentially alleviate the issue.

\subsection{Notations}
We use $\mathcal{G}=(\mathcal{V},\mathcal{E}, \mathbf{X})$ to denote an attributed graph, where $\mathcal{V}=\{v_1,...,v_N\}$ is the set of $N$ nodes, $\mathcal{E} \subseteq \mathcal{V} \times \mathcal{V}$ is the set of edges, and $\mathbf{X}=\{\mathbf{x}_1,...,\mathbf{x}_N\}$ is the set of attributes of $\mathcal{V}$. $\mathbf{A} \in \mathbb{R}^{N \times N}$ is the adjacency matrix of the graph $\mathcal{G}$, where $\mathbf{A}_{ij}=1$ if nodes ${v}_i$ and ${v}_j$ are connected, otherwise $\mathbf{A}_{ij}=0$. In 
our setting, only a limited number of nodes $\mathcal{V}_L=\{v_1,...,v_l\}$ are provided with labels $\mathcal{Y}=\{\mathbf{y}_1,...,\mathbf{y}_l\}$, where $\mathbf{y}_i \in \mathbb{R}^C$ is a one-hot vector of node $v_i$'s label for multi-class classification. Note that the topology of the graph $\mathcal{G}$ could be noisy such as poisoned by adversarial edges or containing inherent noises, which leads to poor performance.
\label{sec:3_1}

\subsection{Basic Design and Inner Working of GNNs}
In this subsection, we briefly introduce the common architecture of graph neural networks (GNNs). 
Generally, GNNs adopt message-passing mechanism to learn node representations, i.e., they update the representation of a node by aggregating the representations of the neighborhood nodes. The updating process of the $k$-th layer in GNNs could be written as:
\begin{equation}
\begin{aligned}
    \mathbf{a}^{(k)}_v & = \text{AGGREGATE}^{(k-1)}(\{\mathbf{h}^{(k-1)}_u: u \in \mathcal{N}(v)\}),
    \label{eq:GNN_a} \\
    \mathbf{h}^{(k)}_{v} & =\text{COMBINE}^{(k)}(\mathbf{h}^{(k-1)}_v, \mathbf{a}_v^{(k)}),
\end{aligned}
\end{equation}
where $\mathbf{h}^{(k)}_v$ is the representation vector of node $v \in \mathcal{V}$ at the $k$-th layer and $\mathcal{N}(v)$ is the set of neighborhoods of $v$. 
During the training of node classification, the representations of labeled nodes are used to give prediction and obtain the training loss to minimize.
With the message-passing mechanism, after $K$-layers of GNN, the node representation of $v_i$ would capture the node features and structure information of the $K$-hop neighborhoods of $v_i$, and thus facilitating downstream tasks. 
In other words, in GNN, \textit{one labeled node would make the $K$-hop neighborhood participate in the training of GNN}, which is one reason that GNNs have great ability in leveraging unlabeled nodes for semi-supervised node classification. 

\subsection{Analysis of GNNs with Sparse Labels}

In this subsection, we conduct preliminary analysis on real-world graphs to show the issues of GNNs when limited labeled nodes are available for training, which paves us a way to design robust GNNs for alleviating the label sparsity issue. The analysis is based on three widely used datasets, i.e., Citeseer~\cite{sen2008collective}, Cora and Cora-ML~\cite{mccallum2000automating}.

Generally, GNNs, such as GCN and GAT, rely on the classification loss of the labeled nodes to learn the parameters,  which is effective when we have adequate labeled nodes. However, when the size of labeled node set $\mathcal{V}_L$ is small and the graph is sparse, only a small portion of nodes would be involved in the training. This may lead to poor performance of GNNs. More specifically, for a $K$-layer GNN, the nodes involved in the training phase include the labeled nodes and the unlabeled nodes within $K$-hop distance of labeled nodes. We usually set $K$ as 2 to 3 because deep GNNs have over-smoothing issue~\cite{li2018deeper}. Since real-world graphs are usually sparse, the $K$-hop neighbors of the labeled nodes would be limited as well. Thus, when $\mathcal{V}_L$ is small, only a small portion of nodes would be involved in training, making GNNs less effective in leveraging unlabeled nodes.

\begin{figure}[t]
\centering
\begin{subfigure}{0.49\columnwidth}
    \centering
    \includegraphics[width=0.9\linewidth]{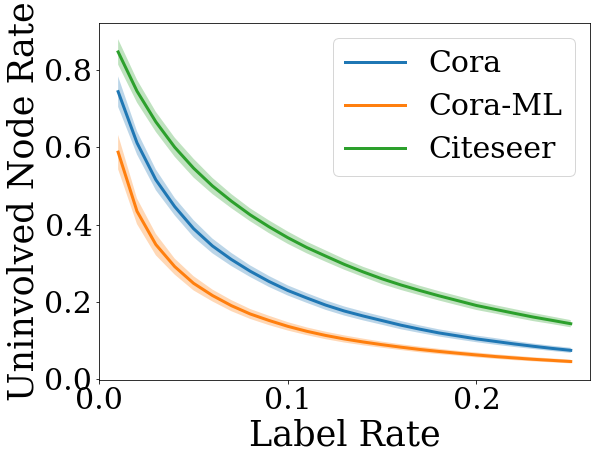} 
    \vskip -0.5em
    \caption{Impacts of label rate}
    \label{fig:1_a}
\end{subfigure}
\begin{subfigure}{0.49\columnwidth}
    \centering
    \includegraphics[width=0.9\linewidth]{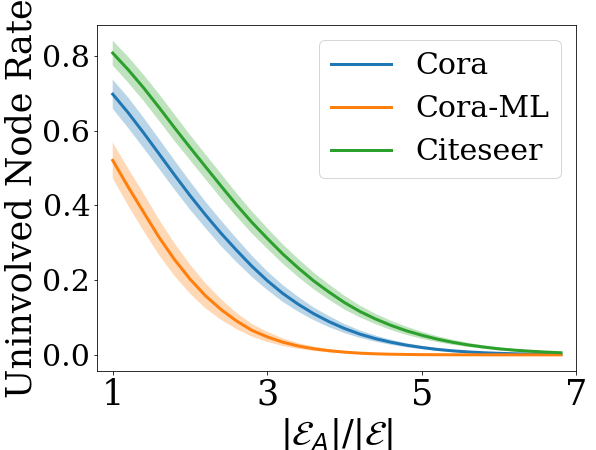} 
    \vskip -0.5em
    \caption{Impacts of graph density}
    \label{fig:1_b}
\end{subfigure}
\vspace{-1.2em}
\caption{The impacts of label rate and density of graph to uninvolved node rate in the training phase. }
\vskip -1.5em
\end{figure}

We analyze how the label rate affects the rates of uninvolved nodes of real-world datasets for a two layer GNN. We vary label rates from 0.01 to 0.25. The average uninvolved node rates and the standard deviations are shown in Fig. \ref{fig:1_a}. From the figure, we observe that (\textbf{i}) when the label rate is high, say above 0.1, most of the nodes are involved in training GNN. The benefit of further increasing label rate is marginal as the 2-hop neighbors of labeled nodes could overlap. This is one reason that GNNs have great ability for semi-supervised node classification with small but adequate amount of labeled nodes, and the increase of labeled nodes can marginally improve the performance; (\textbf{ii}) As the label rate decreases from 0.1, the uninvolved node rate increases significantly, i.e., the majority of nodes are not involved in the training. This indicates that GNNs would have difficulty in handling sparsely labeled graphs.

Although a higher label rate could help to reduce the uninvolved node rate, it can be expensive to obtain more labels~\cite{gallagher2008leveraging}. Thus, we need an alternative approach to effectively use the labels. From the analysis above, one potential solution is to make the graph denser so that one labeled node could have more neighbors to be involved in the training of GNN. To verify it, we randomly add different amount of edges to the three graphs. We denote the number of edges of the new graph as $|\mathcal{E}_A|$ and that of raw graph as $|\mathcal{E}|$. We fix label rate as 0.01. The impact of the graph density on the uninvolved node rate is presented in Fig.~\ref{fig:1_b}. From the figure, we observe that when $|\mathcal{E}_A|/|\mathcal{E}|$ increases from 1 to 3, i.e., we add two times the number of original edges, the uninvoled node rate drops significantly. For example, it drops from 0.8 to around 0.3 on Citeseer. 

As real-world graphs such as social networks have many pairs of nodes who are similar but not connected together, the analysis above shows that it is promising to predict links to densify the graph, which can help the message passing of GNNs to alleviate the issue of limited labeled nodes. In addition, these predicted edges can also be directly used to regularize the predicted labels of unlabeled nodes, i.e., if two nodes are more likely to have a link, they are more likely to have the same labels.

\label{sec:3_3}
\subsection{Problem Definition}
Our preliminary analysis shows that predicting links to densify the graph can potentially alleviate the label sparsity issue.  In addition, the link prediction can potentially down-weight or eliminate noisy edges as noisy edges usually connect nodes with low node attribute similarity.
Therefore, we aim to simultaneously eliminate noisy edges and densify the graph with a link predictor and train a robust GNN on the new graph. The problem is formally defined as:
\newtheorem{problem}{Problem}
\begin{problem}
Given an attributed graph $\mathcal{G}=(\mathcal{V},\mathcal{E}, \mathbf{X})$ with edge set $\mathcal{E}$ might contain a small amount of noisy edges, and a small set of labeled nodes $\mathcal{V}_L \in \mathcal{V}$ with the corresponding labels in $\mathcal{Y}$, simultaneously learn adjacency matrix $\mathbf{S} \in [0,1]^{N \times N}$ which down-weights/removes noisy edges and completes missing links by a link predictor $f_E:(v_i,v_j) \rightarrow \mathbf{S}_{ij}$, and a GNN on the learned graph for node classification, i.e., $f_{\mathcal{G}}:(\mathbf{S}, \mathbf{X}) \rightarrow \hat{\mathcal{Y}}$, where $\mathbf{S}_{ij}$ indicates the weight of edge linking $v_i$ and $v_j$ and $\hat{\mathcal{Y}}$ is the set of predictions for unlabeled nodes.
\end{problem}

\section{Proposed Framework -- RS-GNN}
\label{sec:methodology}
\begin{figure}
    \centering
    \includegraphics[width=0.9\linewidth]{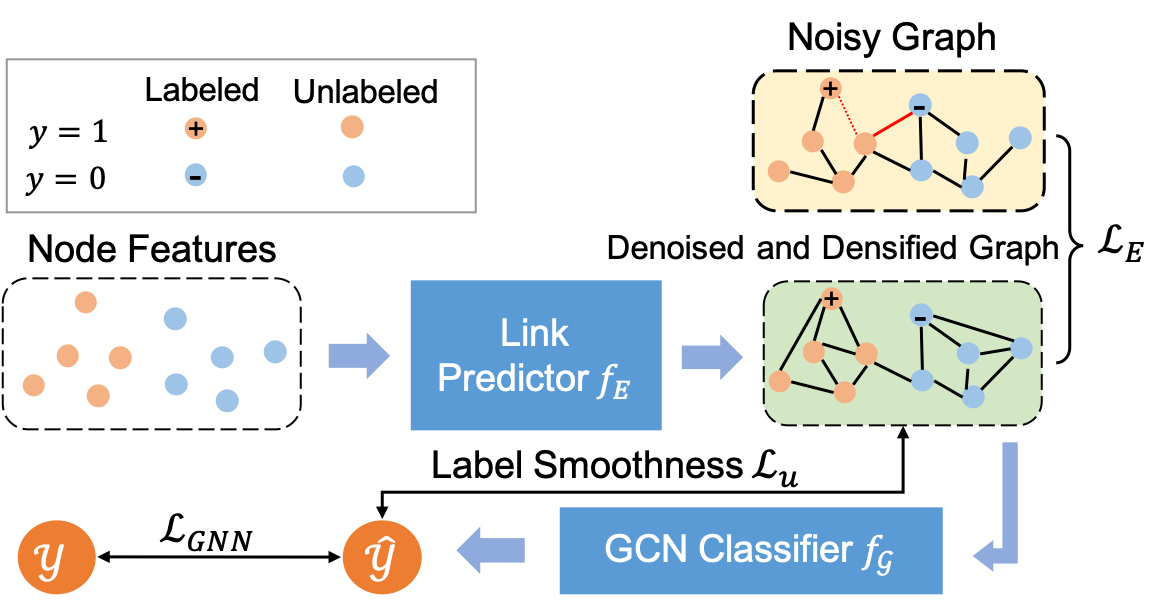}
    \vskip -1em
    \caption{An illustration of the proposed RS-GNN.}
    \label{fig:framework}
    \vskip -1.8em
\end{figure}

In this section, we present the details of the proposed RS-GNN. 
The main challenges are: (i) given the noisy graph, how can we learn a link predictor which can down-weight/eliminate noisy edges and densify the graph; and (ii) how to simultaneously use the  learned graph for node classification. As the graph topology is noisy, we cannot directly apply a GNN on $\mathcal{G}$ to predict edges because the message passing would magnify the negative effects of the noisy edges. Generally, nodes sharing similar features tend to connect to each other; while noisy edges tend to connect nodes of dissimilar nodes. Thus, we propose 
to learn a MLP-based link predictor which predicts links using node attributes. The more similar the node features of two nodes are, the larger weights the link predictor will assign. Thus, the link predictor is able to down-weight or eliminate noisy edges in the initial graph. Meanwhile, the edge predictor can predict missing links to alleviate label sparsity issue. We design a novel feature similarity weighted edge-reconstruction loss to train the link predictor so as to reduce the negative effects of noisy edges on the link predictor.
An illustration of the framework is shown in Figure \ref{fig:framework}, which contains a link predictor $f_E$ and a GCN classifier $f_{\mathcal{G}}$. The link predictor $f_E$ takes node features as input to learn a dense adjacency matrix $\mathbf{S}$, aiming to remove adversarial edges and assign edges that benefit predictions. The GCN classifier $f_{\mathcal{G}}$ takes $\mathbf{S}$ and node features $\mathbf{X}$ to predict the node labels with the node features. Finally, label smoothness constraint based on the predicted edges will be added to the predictions of unlabeled nodes to further alleviate label sparsity issue. Next, we give the details of each component.

\subsection{Link Prediction}
As the given graph contains structural noises and has missing edges, we propose to learn a new graph that down-weights noisy edges to eliminate their negative effects and completes the missing links to facilitate  GNN in dealing noisy graphs with sparse labels. 

\noindent\textbf{Building Link Predictor.} Generally, noisy edges connect two nodes with dissimilar node features; while nodes of similar features are likely to have similar labels and should be connected. Therefore, we propose to predict edge weights and missing edges between nodes using nodes features.
Specifically, for node $v_i$, a MLP takes its node attributes $\mathbf{x}_i$ to learn its node representation as: $\mathbf{z}_i = MLP(\mathbf{x}_i)$.
With the node representations, we predict the weight $w(i,j)$ between $v_i \in \mathcal{V}$ and $v_j \in \mathcal{V}$ as:
\begin{equation}
    w(i,j) = f(\mathbf{z}_i^T \mathbf{z}_j),
    \label{eq:MLP}
\end{equation}
where $f$ is the activation function. For $f$, we use ReLU instead of sigmoid as we find that when the learned adjacency matrix is used as the input of GCN, the use of sigmoid function will lead to gradient vanishing, which is consistent with previous observations~\cite{he2017neural}. Note that we use MLP instead of a GNN as the link predictor because the graph structure is noisy and the message passing of GNN could magnify the negative effects.

\noindent\textbf{Learning Link Predictor.} Our goal is to learn a link predictor which can (i) assign small weights to two nodes of different features so as to eliminate noisy edges; and (ii) assign larger weights to two nodes of similar node features so as to densify the graph to facilitate message passing.
As for many real-world graphs, similar nodes tend to link together and linked nodes usually have high feature similarity. Thus, to learn a good link predictor $f_E$, we utilize the adjacency matrix reconstruction as the loss function. 
Since the graph is sparse, the adjacency matrix $\mathbf{A}$ contains many zero entries. Directly adopting adjacency matrix reconstruction as the loss function would (i) result in poor performance as the link predictor will be biased on predicting missing links; and (ii) require large computational cost as we need to calculate $N^2$ edges. To address this problem, negative sampling~\cite{mikolov2013distributed} is adopted, i.e., for each $v_j \in \mathcal{N}(v_i)$, we randomly sample $Q$ nodes that's not connected to $v_i$ and use them as negative samples. 

However, a small portion of edges in $\mathbf{A}$ are noisy, which might have negative effects in training the predictor. To mitigate the negative effects of noisy edges and to learn a link predictor that can assign lower weights to edges that link dissimilar nodes, we propose to reweight the positive and negative samples based on the feature similarity of two nodes. Specifically, for node  $v_i$ and its positive sample $v_j \in \mathcal{N}(v_i)$, we minimize $\exp(-\frac{\|\mathbf{x}_i-\mathbf{x}_j\|^2}{\sigma^2}) (w(i,j)-1)^2$, where $\sigma$ is the hyperparameter to control the variance of the sample weights. Thus, if the node features of $v_i$ and $v_j$ are similar, $A_{ij}$ is likely to be a clean edge and $\exp(-\frac{\|\mathbf{x}_i-\mathbf{x}_j\|^2}{\sigma^2})$ would be large. Minimizing the loss will force $w(i,j)$ to be close to 1; while if the features are dissimilar, then $A_{ij}$ is likely to be a noisy edge and $\exp(-\frac{\|\mathbf{x}_i-\mathbf{x}_j\|^2}{\sigma^2})$ would be small, thus minimizing the loss will have little effect on $w(i,j)$. Similarly, for $v_i$ and its negative sample $v_n$, we minimize $\exp(\frac{\|\mathbf{x}_i-\mathbf{x}_n\|^2}{\sigma^2}) (w(i,n)-0)^2$. If the node features of $v_i$ and $v_n$ are dissimialr, then $\exp(\frac{\|\mathbf{x}_i-\mathbf{x}_n\|^2}{\sigma^2})$ is large, minimizing the loss would make $w(i,n)$ close to 0 as expected. With the weight defined in this way, the loss for training the link predictor is:
\begin{equation}
\small
\begin{aligned}
    \mathcal{L}_E = \sum_{v_i \in \mathcal{V}} & \sum_{v_j \in \mathcal{N}(v_i)} \Big[\exp(-\frac{\|\mathbf{x}_i-\mathbf{x}_j\|^2}{\sigma^2}) (w(i,j)-1)^2 \\
    & +   \sum_{n=1}^{Q} \cdot \mathbb{E}_{v_n \sim P_n(v_i)} \exp(\frac{\|\mathbf{x}_i-\mathbf{x}_n\|^2}{\sigma^2}) (w(i,n)-0)^2\big],
    \label{eq:edge}
\end{aligned}
\end{equation}
where $P_n(v_i)$ is the distribution of sampling negative nodes for $v_i$, which is a uniform distribution. With the loss function Eq.(\ref{eq:edge}), the link predictor would be able to downweight the noisy edges and densify the graph to facilitate the learning of robust GNN on noisy graph with limited labels. 

\noindent\textbf{Graph Denoising and Densification.} 
With the link predictor, we could apply the learned weights to the existing edges and drop edges whose predicted weights are small to eliminate the negative effects of noisy/adversarial edges. Moreover, to increase the involvement of unlabeled nodes to facilitate the message passing of GNNs, we also link nodes that have large weights predicted by the link predictor. However, if we predict weights of all pairs of nodes, the computation cost will be very large because we will train a link predictor and a GNN classifier end-to-end as shown in Sec.~\ref{sec:4_4}, which means we need to do prediction in each iteration. To save the computational cost, for each node $v_i$, we first construct a candidate subset $\mathcal{S}(v_i)$, which contains $K$ nodes having the largest cosine similarities with $v_i$ in the raw feature space $\mathbf{X}$. Note that this only needs to be done once. Since nodes not in $\mathcal{S}(v_i)$ are not likely to be connected with $v_i$, we only need to compute weights between $v_i$ and $\mathcal{S}(v_i)$. The whole process of obtaining a clean and dense adjacency matrix $\mathbf{S}$ could be formally stated as:
\begin{equation}
    \mathbf{S}_{ij} = \left\{ \begin{array}{ll}
         w(i,j) & \mbox{if $w(i,j) > T_l$ and $v_j \in \mathcal{N}(v_i) \cup \mathcal{S}(v_i)$} ;\\
        0 & \mbox{else},\end{array} \right.
        \label{eq:generate_graph}
\end{equation}
where $\mathcal{N}(v_i)$ are neighbors of $v_i$ in the noisy graph, and $T_l$ is a threshold to determine whether we should keep/add the edge.  With the above operation, those noisy edges would be assigned smaller weights or even dropped, which mitigate the negative effects of noisy edges. Meanwhile, more edges are introduced to facilitate the message passing of GNNs during training.

\subsection{GNN for Node Classification}
With the learned adjacency matrix $\mathbf{S}$, we can apply GNNs to learn the node representation as $\mathbf{H} = GNN(\mathbf{S}, \mathbf{X})$. 
Note that the proposed framework is a flexible framework which can facilitate various GNNs such as GAT~\cite{velivckovic2017graph} and GIN~\cite{xu2018powerful}. With the node representation, the label of node $v_i$ can be predicted as
$\hat{y}_i = softmax(\mathbf{h}_i)$, where $\mathbf{h}_i$ is the representation of node $v_i$. Then, the training loss is:
\begin{equation}
\small
    \mathcal{L}_{GNN} = \sum_{v_i \in \mathcal{V}_L} l(\mathbf{\hat{y}}_i, \mathbf{y}_i)
    \label{eq:GNN_dense}
\end{equation}
where $l(\mathbf{\hat{y}}_i, \mathbf{y}_i)$ is the cross entropy between $\hat{y}_i$ and $\mathbf{y}_i$. Since $\mathbf{S}$ is denser than the original graph, more unlabeled nodes are involved in the training even with limited amount of labeled nodes, thus making the propagation of information more efficient.

\subsection{Label Smoothness on Unlabeled Nodes}
Though the dense graph $\mathbf{S}$ can help to include more unlabeled nodes in the loss function, their information is propagated through the message-passing mechanism instead of being directly used in the training loss. To further alleviate the issue of limited labeled nodes, we  propose to adopt the predicted weighted edges for label smoothness regularization. The basic idea is the larger weights of an edge $S_{ij}$ is, the more likely that $v_i$ and $v_j$ have the same label~\cite{wang2019knowledge}.
Thus, for an unlabeled node $v_i$, if its edge weight with node $v_j$ is larger than a threshhold $T_h$, i.e., $S_{ij} > T_h$, we want their predicted labels to be similar with each other. This can be formally written as
\begin{equation}
    \mathcal{L}_u = \sum_{v_i \in \mathcal{V}_u}\sum_{v_j \in \mathcal{V}} \mathbf{T}_{ij} \|\mathbf{\hat{y}}_i-\mathbf{\hat{y}}_j\|^2,
\end{equation}
where $\mathcal{V}_u$ denotes the set of unlabeled nodes, $\mathbf{\hat{y}}_i$ and $\mathbf{\hat{y}}_j$ represent the predictions of node $v_i \in \mathcal{V}_u$ and $v_j \in \mathcal{V}$, respectively. $\mathbf{T}_{ij}=\mathbf{S}_{ij}$ if $\mathbf{S}_{ij}>T_h$; otherwise 0.
In this way, we explicitly smooth the predicted labels between unlabeled nodes and nodes that are similar to them. By including $\mathbf{T}_{ij}$ in $\mathcal{L}_u$, edge weights are also considered. 

\subsection{Final Objective Function of RS-GNN} \label{sec:4_4}
With the link predictor denoising and densifying the graph with the supervision from $\mathbf{A}$, the GNN adopting the learned graph for label prediction and the label smoothness regularization from the generated graph, the final loss function can be written as
\begin{equation}
    \mathop{\arg \min}_{\theta_E,\theta_{\mathcal{G}}} \mathcal{L}_{GNN} + \alpha \mathcal{L}_E + \beta \mathcal{L}_u,
    \label{eq:final}
\end{equation}
where $\theta_E$ and $\theta_{\mathcal{G}}$ are parameters of link predictor $f_E$ and GNN classifier $f_\mathcal{G}$, respectively. $\alpha$ and $\beta$ are hyperparameters to balance the contributions of reconstructing the adjacency matrix with $f_E$ and label smoothness regularization. The proposed framework is an end-to-end framework where we simultaneously learn the link predictor and utilize the predicted edges for training a robust GNN to alleviate the noisy graph and limited labeled nodes issues. The training algorithm is shown in the supplementary material.



\section{Experiments}

\label{Sec:experiments}

\begin{table*}[t]
    \small
    \centering
    \caption{Node classification performance (Accuracy(\%)$\pm$Std) on various types of noisy graphs}
    \vskip -1.5em
    \begin{tabularx}{0.985\textwidth}{|p{0.05\textwidth}|p{0.14\textwidth}|CC>{\centering\arraybackslash}p{0.1\linewidth}C>{\centering\arraybackslash}p{0.1\linewidth}>{\centering\arraybackslash}p{0.08\linewidth}CC|}
    \hline
    Dataset & Graph & GCN & SuperGAT &Self-Training & RGCN & GCN-jaccard & GCN-SVD & Pro-GNN & Ours \\
    \hline
    
    \multirow{4}{*}{Cora}
        &Raw Graph            & 65.5 $\pm 0.5$& 69.0 $\pm 1.7$ & 67.9 $\pm 0.9$ & 63.0 $\pm 0.7$ &65.7 $\pm 0.6$ & 62.9 $\pm 1.1$  & 65.9 $\pm 1.3$ & \textbf{75.3} $\pm \textbf{0.6}$\\
        &Random Noise        & 59.2 $\pm 0.7$ & 58.8 $\pm 0.4$ & 63.1 $\pm 0.5$ &51.5 $\pm 0.7$ & 57.8 $\pm 1.4$ & 51.5 $\pm 0.7$ & 56.1 $\pm 3.0$ & \textbf{71.8} $\pm \textbf{1.5}$\\
        &Non-Targeted Attack  & 26.8 $\pm 2.5$ & 41.5 $\pm 1.6$ & 29.6 $\pm 0.4$ &30.4 $\pm 1.0$ & 48.3 $\pm2.0$ & 37.1 $\pm 1.4$ & 41.7 $\pm 5.7$& \textbf{70.8} $\pm \textbf{0.7}$  \\
        &Targeted Attack      & 45.3 $\pm 1.2$& 44.4 $\pm 1.3$ &46.7 $\pm 2.1$ &40.3 $\pm 1.0$ & 49.5 $\pm 1.0$ & 44.8 $\pm 0.7$ & 49.7 $\pm 0.9$ & \textbf{67.8} $\pm \textbf{1.2}$ \\

    \hline
    \multirow{4}{*}{Cora-ML}
        &Raw Graph         & 72.4 $\pm 0.8$ & 73.8 $\pm 1.4$ & 72.7 $\pm 1.4$ & 72.9 $\pm 0.7$ & 71.0 $\pm 1.2 $ & 71.1 $\pm 1.0$ & 62.0 $\pm 1.5$ & \textbf{75.6} $\pm \textbf{0.4}$\\
        &Random Noise       & 62.3 $\pm 0.6$ & 63.7 $\pm 0.9$ & 62.8 $\pm 1.3$ & 61.4 $\pm 1.1$ & 61.3 $\pm 0.5$ & 62.6 $\pm 0.6$ & 57.1 $\pm 2.1$ & \textbf{72.9} $\pm \textbf{0.7}$\\
        &Non-Targeted Attack & 13.2 $\pm 1.4$ & 18.6 $\pm 1.5$ & 15.0 $\pm 0.7$ & 11.0 $\pm 1.0$ & 48.9 $\pm 5.3$ & 16.3 $\pm 0.6$ & 18.2 $\pm 2.4$ & \textbf{73.2} $ \pm \textbf{1.2}$ \\
        &Targeted Attack     & 55.7 $\pm 0.7$ & 56.5 $\pm 1.7$ & 57.7 $\pm 1.2$ & 54.6 $\pm 0.6$ & 61.2 $\pm 0.9$ & 53.0 $\pm 0.8$ & 55.1 $\pm 1.6$ & \textbf{70.8} $\pm \textbf{0.7}$ \\

    \hline
    \multirow{4}{*}{Citeseer}
        &Raw Graph           & 64.8 $\pm 1.4$ & 64.2 $\pm 1.7$ & 65.7 $\pm 1.1$ & 56.6 $\pm 1.2$ & 62.2 $\pm 2.0$ & 61.3 $\pm 2.0$ & 60.6 $\pm 2.0 $  & \textbf{71.2} $\pm \textbf{1.4}$\\
        &Random Noise       & 57.0 $\pm 1.2$ & 54.6 $\pm 1.3$ & 58.7 $\pm 2.1$ & 48.2 $\pm 1.2$ & 61.1 $\pm 2.8$ & 48.3 $\pm 1.6$ & 54.4 $\pm 2.6$ & \textbf{68.8} $\pm \textbf{1.5}$\\
        &Non-Targeted Attack & 26.6 $\pm 2.5$ & 42.3 $\pm 2.6$ & 28.8 $\pm 2.7$ &26.6 $\pm 1.1$ & 57.9 $\pm 2.7$ & 41.7 $\pm 1.6$ & 41.6 $\pm 3.1$ & \textbf{68.0} $\pm \textbf{0.4}$ \\
        &Targeted Attack     & 43.9 $\pm 1.7$ & 42.9 $\pm 0.4$ & 47.6 $\pm 1.2$ &35.3 $\pm 1.5$ & 52.5 $\pm 2.3$ & 40.5 $\pm 0.7$ & 48.1 $\pm 1.6$ & \textbf{67.2} $\pm \textbf{1.3}$\\
    \hline
    \multirow{4}{*}{Pubmed}
    & Raw Graph          & 85.9 $\pm 0.1$ & 86.0 $\pm 1.2$ & 86.1 $\pm 0.2$ & 85.1 $\pm 0.1$ & 86.0 $\pm 0.1$  & 83.0 $\pm 0.1$ & 86.1 $\pm 0.1$ & \textbf{86.9} $\pm \textbf{0.1}$  \\
    & Random Noise & 80.5 $\pm 0.1$ & 79.8 $\pm 0.1$ & 81.2 $\pm 0.2$ & 79.7 $\pm 0.1$ &  83.0 $\pm 0.1$  & 82.0 $\pm 0.1$ &  85.1 $\pm 0.2$ &  \textbf{86.4} $\pm \textbf{0.1}$\\
    & Non-Targeted Attack & 73.7 $\pm 0.2$ & 73.8 $\pm 0.2$ & 73.5 $\pm 0.3$ & 73.8 $\pm 0.3$ & 84.4 $\pm 0.1$ & 83.0 $\pm 0.1$ & 86.0 $\pm$ 0.1 & \textbf{86.3} $\pm \textbf{0.1}$ \\
    & Targeted Attack & 76.5 $\pm 0.1$ & 75.6 $\pm 0.1$ & 76.8 $\pm 0.2$ & 76.2 $\pm 0.2$  & 82.7 $\pm 0.2$ &78.1 $\pm 1.3$ & 79.1 $\pm 0.1$ & \textbf{84.3} $\pm \textbf{0.2}$ \\
    \hline
    \end{tabularx}
    
    \label{tab:results}
    \vskip -1.2em
\end{table*}

In this section, we evaluate the proposed RS-GNN on noisy graphs with limited labels to answer the following research questions:
\begin{itemize}[leftmargin=*]
    \item \textbf{RQ1} How robust is the proposed framework on various types of noisy graphs with limited labeled nodes?
    \item \textbf{RQ2} How does the proposed framework perform under various label rates and graph sparsity levels?
    \item \textbf{RQ3} What are the contributions of link predictor and label smoothness regularization from predicted edges on RS-GNN?
\end{itemize}
\subsection{Experimental Settings}
\label{Sec:ex_settings}

\subsubsection{Datasets} 
\label{Sec:datasets}
For a fair comparison, we conduct experiments on four widely used benchmark datasets, i.e., Cora, Cora-ML, Citeseer and Pubmed~\cite{sen2008collective}.
The statistics of the datasets are presented in the Table \ref{tab:dataset} in Appendix. Note that the split of validation and testing on all datasets are the same as described in the cited papers to keep consistence. For the training set, we randomly sample 1\% of nodes as the labeled set for Cora, Cora-ML and Citeseer. For Pubmed, we randomly sample 10\% of nodes to compose the labeled set. The training node set doesn't overlap with the validation and test sets. 

\subsubsection{Noisy Graphs}
To show RS-GNN is robust to various structural noises, we evaluate RS-GNN on the following types of noises:
\begin{itemize}[leftmargin=*]
    \item \textbf{Raw Graphs}: They are the original graphs of the benchmark datasets which may contain inherent structural noise.
    \item \textbf{Random Noise}: We randomly inject fake edges and remove normal edges to add random noise to graphs.
    \item \textbf{Non-Targeted Attack}: 
    We adopt \textit{metattack}~\cite{zugner2019adversarial} to poison the graph structures by adding and removing edges, which aims to reduces the overall performance of GNNs on the whole graph. 
    \item \textbf{Targeted Attack}: It aims to lead the GNN to misclassify target nodes. Following~\cite{tang2020transferring}, we randomly select 15\% nodes as target nodes and apply \textit{nettack}~\cite{zugner2018adversarial} to perturb the graph structure. 

\end{itemize}

\subsubsection{Baselines} We compare RS-GNN with the representative and state-of-the-art GNNs, and robust GNNs against adversarial attacks:
\begin{itemize}[leftmargin=*]
    \item \textbf{GCN}~\cite{kipf2016semi}: GCN is a representative GNN which defines Graph convolution with spectral analysis.
     \item \textbf{SuperGAT}~\cite{kim2021find}: This extends GAT~\cite{velivckovic2017graph} with self-supervised learning. Edge prediction is deployed as the pretext task to guide the learning of attention to facilitate the message-passing.
    \item \textbf{Self-Training}~\cite{li2018deeper}: This is a self-supervised learning method. A GCN is firstly trained on given labels. Then, confident pseudo labels would be added to the label set to improve the GCN.
    \item \textbf{RGCN}~\cite{zhu2019robust}: It uses Gaussian distributions as representations to absorb the effects of adversarial edges. 
    \item \textbf{GCN-jaccard}~\cite{wu2019adversarial}: GCN-Jaccard eliminates edges that connect nodes with low Jaccard similarity, then apply GCN on the graph.
    \item \textbf{GCN-SVD}~\cite{entezari2020all}: This preprocessing method is based on low rank assumption. Low-rank approximation of the perturbed graph is used to train GNNs against adversarial attacks.
    \item \textbf{Pro-GNN}~\cite{jin2020graph}: It applies low-rank and sparsity constraints to learn a clean graph structure close to the noisy graph structure. 
\end{itemize}
For all the baselines, we use the implementation from the repository DeepRobust~\cite{li2020deeprobust}. All the hyperparameters of the baselines are tuned on the validation set to make a fair comparison with RS-GNN.

\subsubsection{Implementation Details}
\label{sec:implementation}
\textit{Each experiment is conducted 5 times} and average results with standard deviations are reported. The hyperparameters are tuned based on the performance of validation set. More specifically, for RS-GNN, we vary $\alpha$ as  \{0.003, 0.03, 0.3, 3, 30 \}, and $\beta$ as \{0.01, 0.03, 0.1, 0.3, 1\}. For all experiments, $T_l$, $T_h$, $\sigma$, and $Q$ are fixed as 0.1, 0.8, 100, and 50, respectively. $K$ is set as 100, 300, 400 and 10 for Cora, Cora-ML, Citeseer and Pubmed, respectively. More details about the hyperparameters sensitivity is discussed in Sec. \ref{Sec:para_analysis}.
A one-hidden layer MLP with 64 filters is applied as the link predictor. We use GCN as the backbone of RS-GNN. Various GNNs can be used in RS-GNN and we leave it as a future work.

\begin{figure}[t]
\centering
\begin{subfigure}{0.49\columnwidth}
    \centering
    \includegraphics[width=0.85\linewidth]{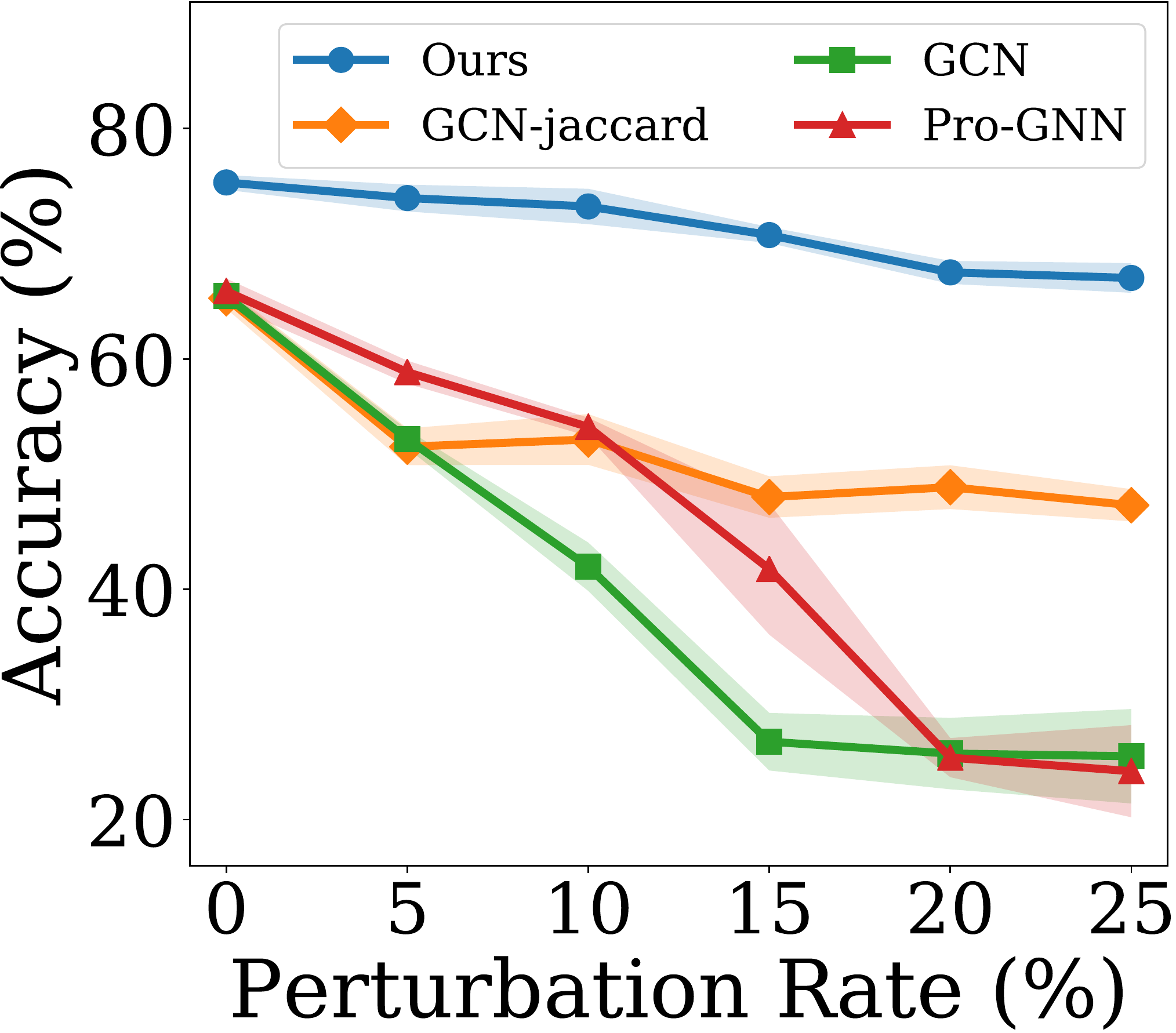} 
    \vskip -0.5em
    \caption{Metattack}
\end{subfigure}
\begin{subfigure}{0.49\columnwidth}
    \centering
    \includegraphics[width=0.85\linewidth]{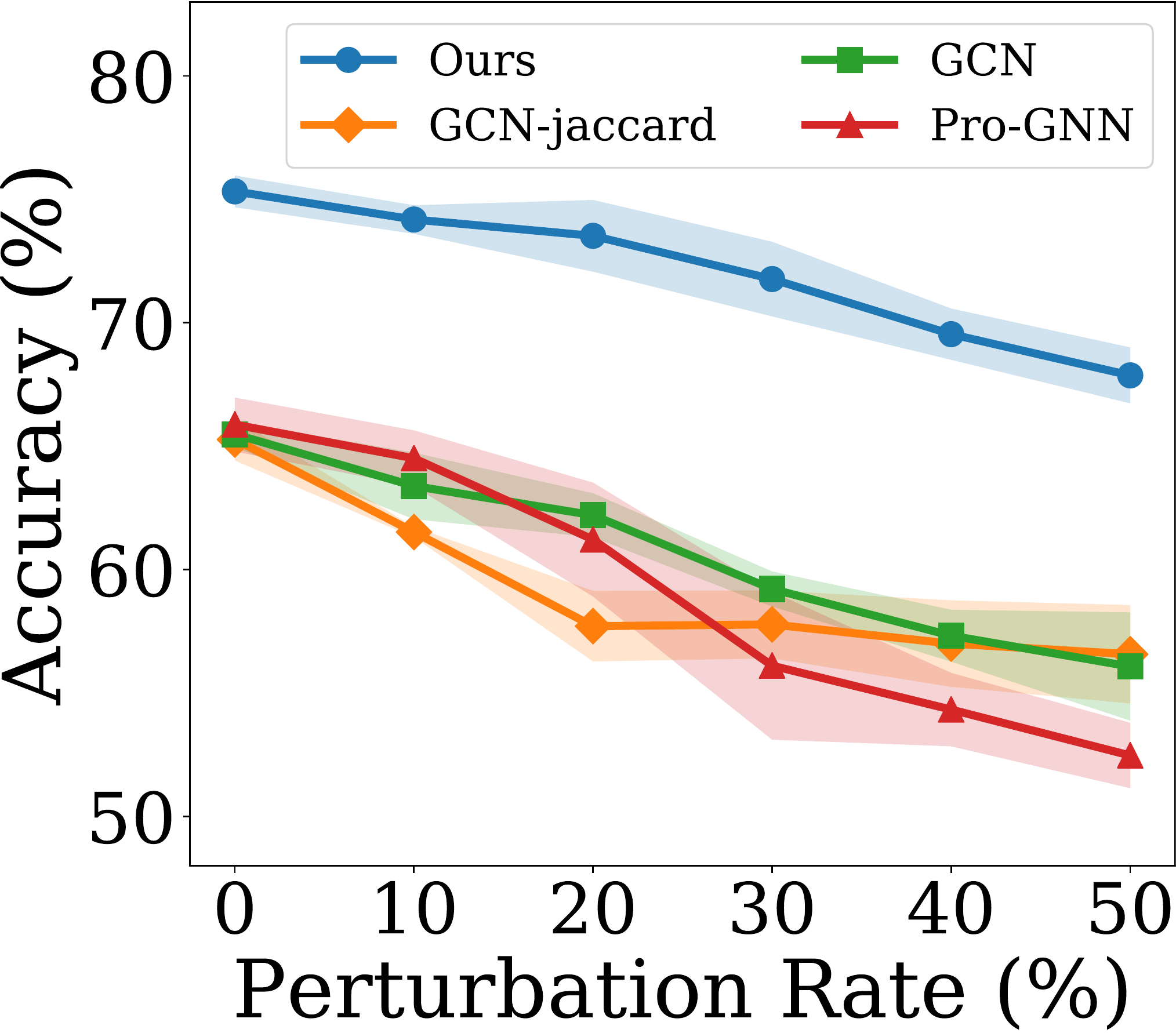} 
    \vskip -0.5em
    \caption{Random Noise}
\end{subfigure}
\vspace{-1.2em}
\caption{Robustness under different Ptb rates on Cora.  }
\label{fig:ptb}
\vskip -1.5em
\end{figure}

\subsection{Performance on Noisy Graphs}
To answer \textbf{RQ1}, we first compare RS-GNN with the baselines on various noisy graphs. We then evaluate the performance of RS-GNN on the graphs with different levels of structural noise.

\subsubsection{Comparisons with baselines}
We conduct experiments on four types of noisy graphs, i.e., raw graphs, graphs with random noise, non-targeted attack perturbed graphs and targeted attack perturbed graphs. The perturbation rate of non-targeted attack and targeted attack is 0.15. The perturbation rate of random noise is set as 0.3. Since we focus on noisy graph with sparse labels, we set the label rates as 0.01 for Cora, Cora-ML, Citeseer and 0.1 for Pubmed. The results are reported in Table \ref{tab:results}, where we can observe:

\begin{itemize}[leftmargin=*]
    \item With limited labeled nodes, GCN even hardly performs well on raw graph, which indicates the necessity of investigating method to address the challenge of sparsely labeled graphs. Though recent GNNs such as SuperGAT and Self-Training can improve the performance with self-supervised learning, our RS-GNN still outperforms them by a large margin. This shows the effectiveness of graph densification in dealing with sparsely labeled graphs.
    \item The structural noise further degrades the performance of GCN, but its impact to RS-GNN is negligible. RS-GNN achieves better results than the state-of-the-art robust GNNs. This indicates RS-GNN could eliminate the effects of the noisy edges.
    \item Compared with the preprocessing methods and Pro-GNN, RS-GNN achieves higher accuracy on the sparsely labeled graphs perturbed by attack methods. 
    This is because the baselines only focus on eliminating potential noisy edges, which will even result in less involvement of unlabeled nodes. 
    By contrast, RS-GNN can down-weights/removes the adversarial edges to defend the adversarial attacks and densify the graph to facilitate the message passing for predictions of unlabeled nodes.
    
\end{itemize}

\subsubsection{Robustness Under Different Ptb Rates } 
To show that RS-GNN is resistant to different levels of structural noise, we vary the perturbation rate as $\{0\%, 5\%, 10\%, \dots, 25\%\}$ and compare the performance of RS-GNN with the most effective baselines. The label rate is fixed as 0.01. Since we have similar observations on other datasets, we only report the average accuracy and standard deviation on Cora in Figure \ref{fig:ptb}. From the figure, we make following observations:
\begin{itemize}[leftmargin=*]
    \item As the perturbation rate increases, the performance of all the baselines drop significantly, which is as expected. Though the performance of RS-GNN also drops, it is much stable and consistently outperforms the baselines, which shows the robustness of RS-GNN against various levels of attacks and random noise; and 
    \item  Compared with GCN, RS-GNN uses GCN as backbone but significantly outperforms GCN, especially when the perturbation rate is large, which shows the effectiveness of eliminating the effects of noisy edges and densifying the graph to benefit the predictions given limited labels. 
\end{itemize}

\begin{figure}[t]
\centering
\begin{subfigure}{0.49\columnwidth}
    \centering
    \includegraphics[width=0.85\linewidth]{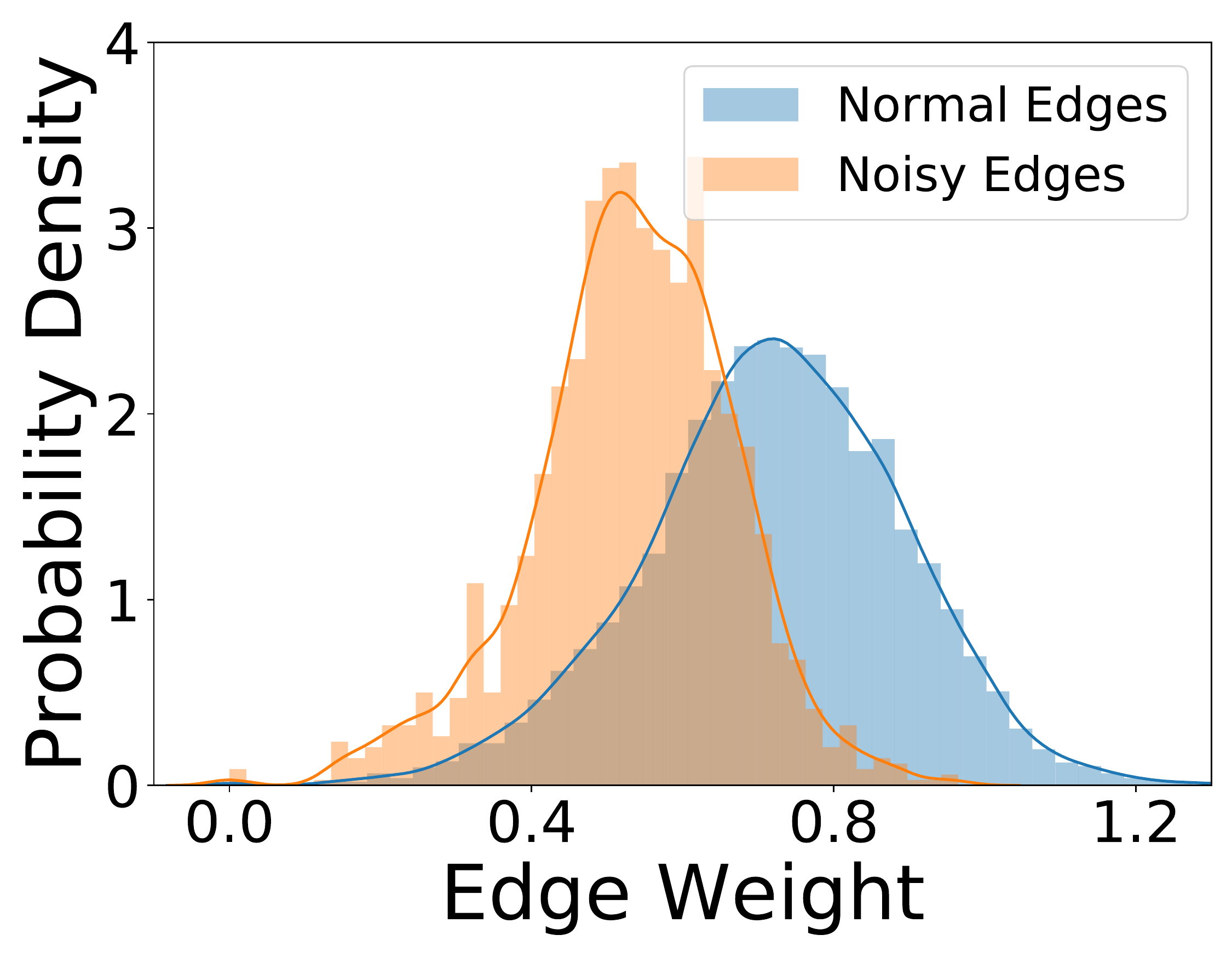} 
    \vskip -0.8em
    \caption{Cora}
\end{subfigure}
\begin{subfigure}{0.49\columnwidth}
    \centering
    \includegraphics[width=0.85\linewidth]{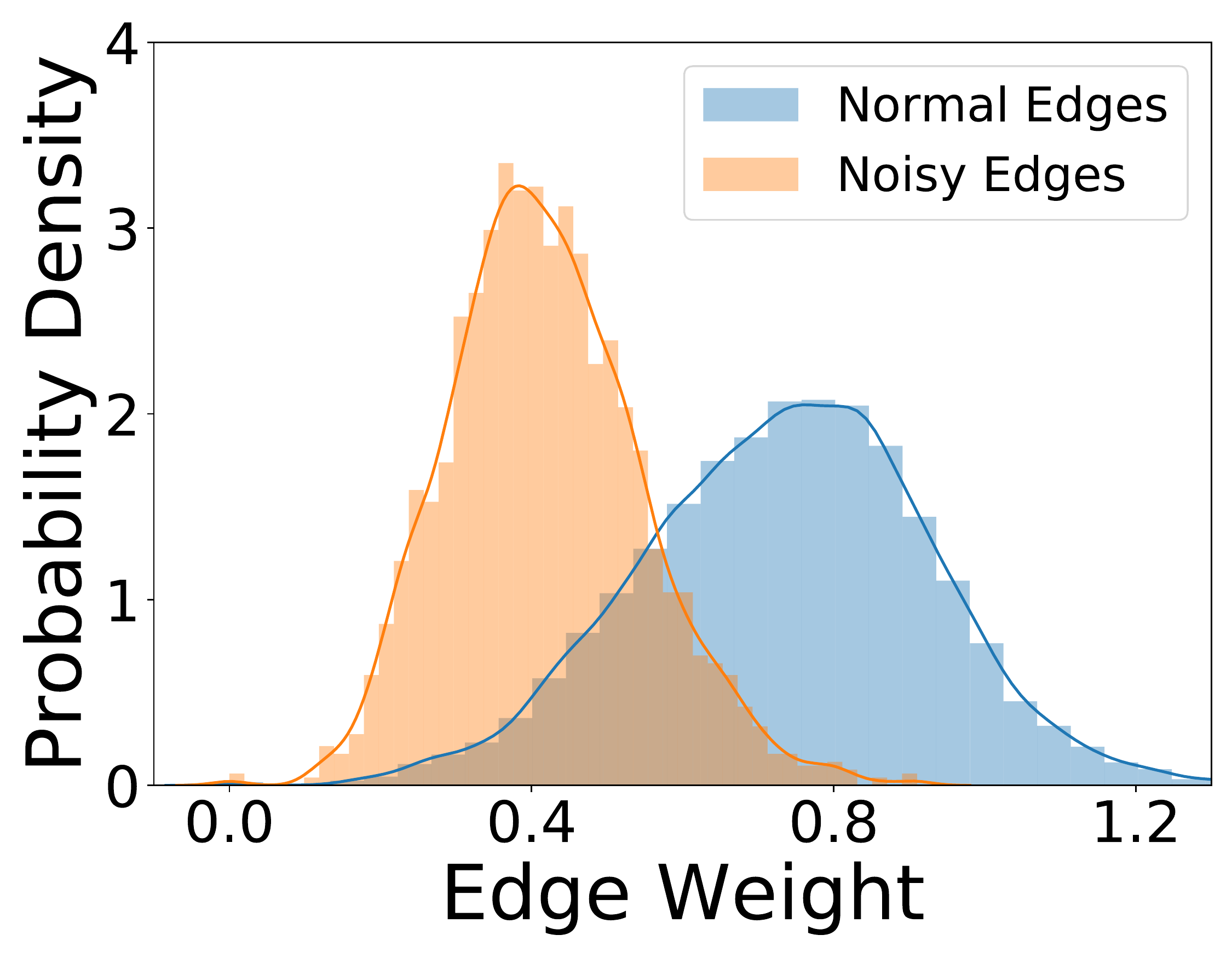} 
    \vskip -0.8em
    \caption{CoraML}
\end{subfigure}
\vspace{-1.5em}
\caption{ Distributions of the weights of normal and noisy edges on the generated graph.}
\label{fig:weight}
\vskip -1em
\end{figure}

\subsection{Analysis of the Learned Graph} 
To demonstrate that RS-GNN could alleviate negative effects of noisy edges by downweighting the noisy edges, we investigate the distribution of the learned edge weights $\mathbf{S}_{ij}$ of normal and noisy edges in this subsection. The edge weight  distributions of graphs perturbed by random noise with 30\% perturbation rate on Cora and Cora-ML are shown in Fig.~\ref{fig:weight}.  From this figure, we observe: (\textbf{i}) The weights of noisy edges are significantly lower than the weights of normal edges, which indicates RS-GNN manages to reduce the effects of noisy edges for robust GNN; and (\textbf{ii}) Although most normal edges have higher weights, some of their weights are very low, which implies inherent noise exists in the graph and RS-GNN is able to get rid of such inherent structural noise. 

We also provide more details about the number of involved unlabeled nodes with the learned graph in Appendix~\ref{sec:app_graph}, which proves RS-GNN can enhance the involvement of unlabeled nodes.

\begin{figure}[t]
\centering
\begin{subfigure}{0.49\columnwidth}
    \centering
    \includegraphics[width=0.9\linewidth]{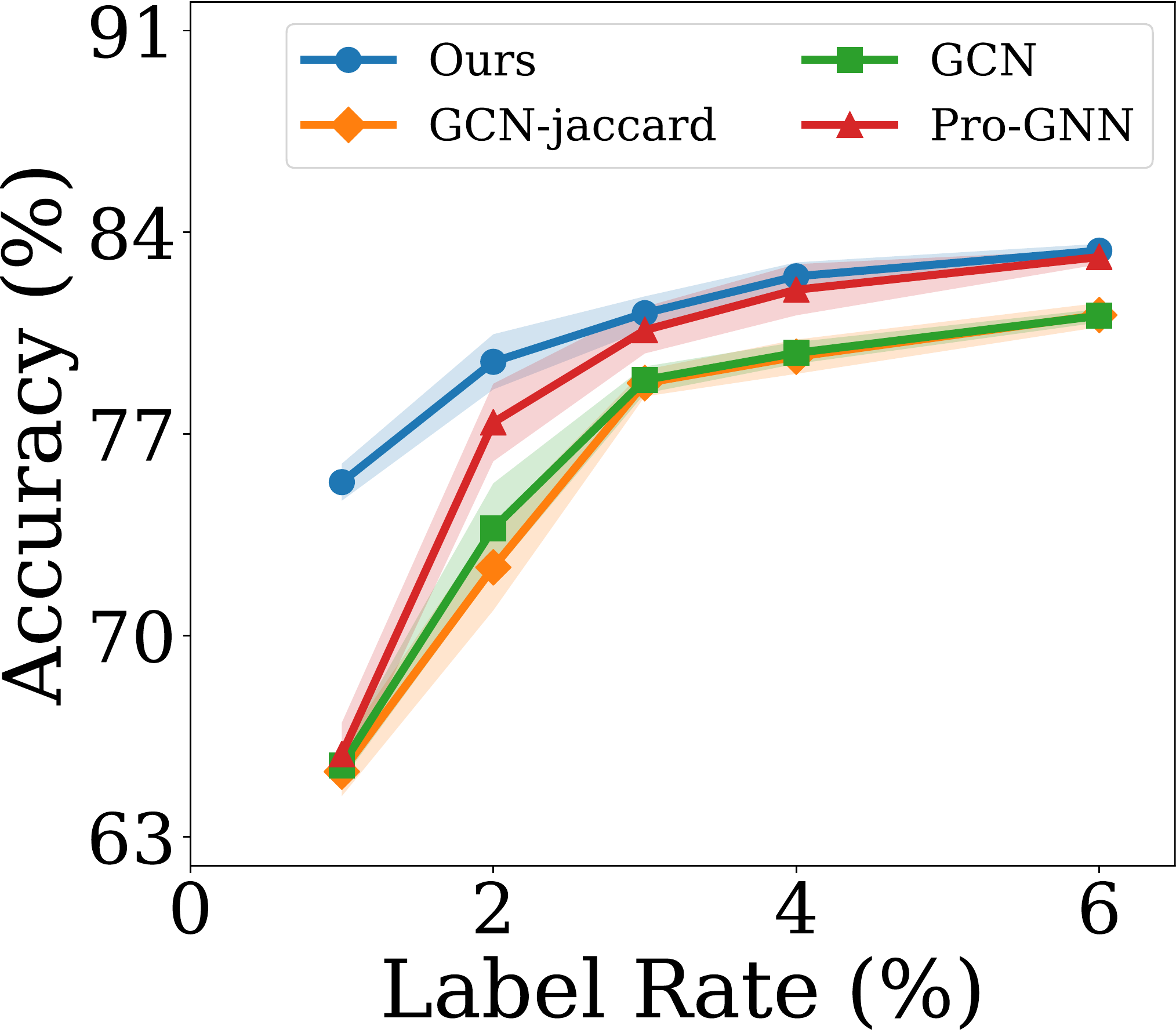} 
    \vskip -0.5em
    \caption{Raw Graph}
\end{subfigure}
\begin{subfigure}{0.49\columnwidth}
    \centering
    \includegraphics[width=0.9\linewidth]{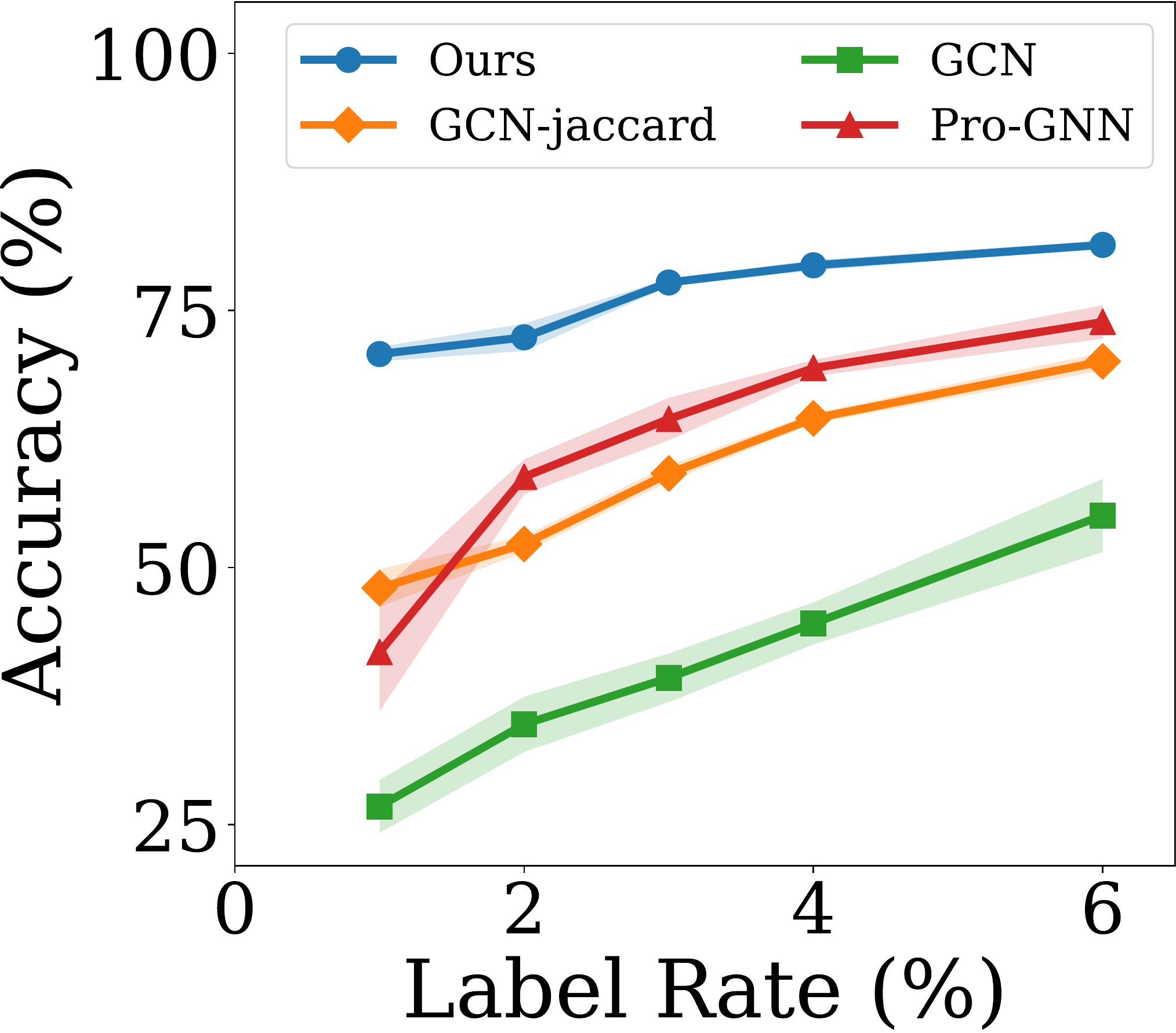} 
    \vskip -0.5em
    \caption{Metattack with 15\% Ptb}
\end{subfigure}
\vspace{-1.2em}
\caption{Performance on Cora with different label rates.  }
\label{fig:label_rate}
\vskip -0.8em
\end{figure}

\subsection{Impacts of Label Rate and Graph Sparsity}

To answer \textbf{RQ2}, we study the impacts of the number of labeled nodes and sparsity of the graph by varying the label rate and edge rate of the graph. The hyperparameters are selected with the process described in Sec. \ref{sec:implementation}. Each experiment is conducted 5 times and average accuracy with standard deviation are reported.

\subsubsection{Impacts of Label Rate} We vary label rates as \{0.01, 0.02,\dots, 0.06\}. Experiments are conducted on raw graphs and graphs perturbed by \textit{mettack} to study the effectiveness of RS-GNN under various label rates. The results on Cora are shown in Fig.~\ref{fig:label_rate}. We have similar observations on other datasets. From Fig.~\ref{fig:label_rate}, we observe:
\begin{itemize}[leftmargin=*]
    \item Generally, as the increase of label rate, the performances of all the methods increase, which is as expected.
    \item For the raw graph, though RS-GNN consistently outperforms the baselines, as the label rate increases, the improvement of RS-GNN becomes marginal. This is because the raw graph doesn't contain much noise. Thus, as label rate increases to 6\%, there are already adequate labels. Since higher label rates would result in more unlabeled nodes involving in the training, the effects of densifying graphs and label smoothness become less significant; 
    \item For the metattack graph, as the label rate increases, RS-GNN still significantly outperforms baselines. That's because the training graph contains a lot of adversarial edges. Though we have enough training labels, the adversarial edges can still contaminate the message passing of GNNs. But RS-GNN can eliminate noisy edges and densify the graph, thus having better results.
\end{itemize}

\subsubsection{Impacts of Graph Sparsity} As RS-GNN can generate dense graphs, it should have the ability to handle sparse graphs. Thus, we randomly select $x\%$ edges from the raw graph to build graphs of different sparsity levels. We vary edge rate $x\%$ from 20\% to 100\% with a step of 40\%. Since we are interested in how the sparsity of the graph could affect RS-GNN in generating dense graphs, we only focus on the performance on raw graphs. 
The average results of 5 runs on Citeseer are reported in Table~\ref{tab:sparsity}. From the table, we have the following observations:
\begin{itemize}[leftmargin=*]
    \item As the edge rate decreases, the performance of all the methods decrease, which is because message-passing of GNNs becomes ineffective on very sparse graphs;
    \item RS-GNN consistently outperforms the baselines. In particular, when the graph becomes more sparse, the improvement of RS-GNN over the baselines becomes larger. For example, the improvement of RS-GNN over GCN on Citeseer is 6.4\% when Edge Rate is 100\%, and becomes 9.2\% when Edge Rate is 20\%, which shows the importance of generating edges for densifying the graph and smoothing predictions with the learned graph. 
\end{itemize}

\begin{table}[t]
    \small
    \centering
    \caption{Accuracy (\%) on Citeseer in different sparsity levels. }
    \vskip-1.5em
    \begin{tabularx}{0.96\linewidth}{>{\centering\arraybackslash}p{0.20\linewidth}CCC}
    \toprule
    Edge Rate (\%) & GCN & Pro-GNN & RS-GNN\\
    \midrule

    20 & 54.5 $\pm 1.2$ & 55.2 $\pm 1.6$ & \textbf{63.7} $\pm \textbf{2.2}$\\
    60 & 58.7 $\pm 1.8$ & 58.3 $\pm 2.4$ &
    \textbf{69.8} $\pm \textbf{1.1}$\\
    100 & 64.8 $\pm 1.4$ & 60.6 $\pm 2.0$ &
    \textbf{71.2} $\pm \textbf{1.4}$\\
    \bottomrule
    \end{tabularx}
    \label{tab:sparsity}
    \vskip -1.em
\end{table}


\begin{figure}[h]
\centering
\begin{subfigure}{0.49\columnwidth}
    \centering
    \includegraphics[width=0.85\linewidth]{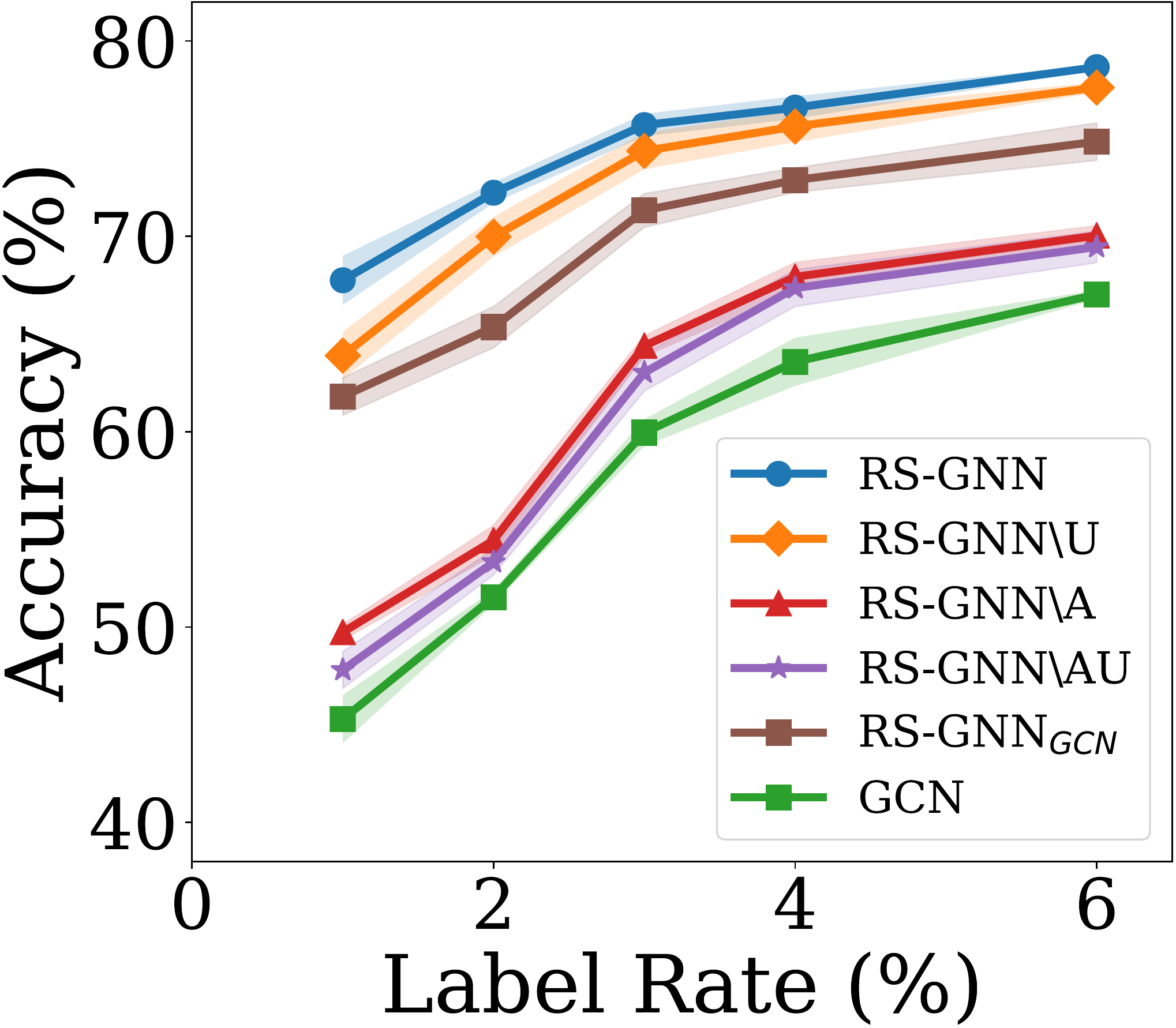} 
    \vskip -0.5em
    \caption{Nettack}
    \label{fig:abla_neta}
\end{subfigure}
\begin{subfigure}{0.49\columnwidth}
    \centering
    \includegraphics[width=0.85\linewidth]{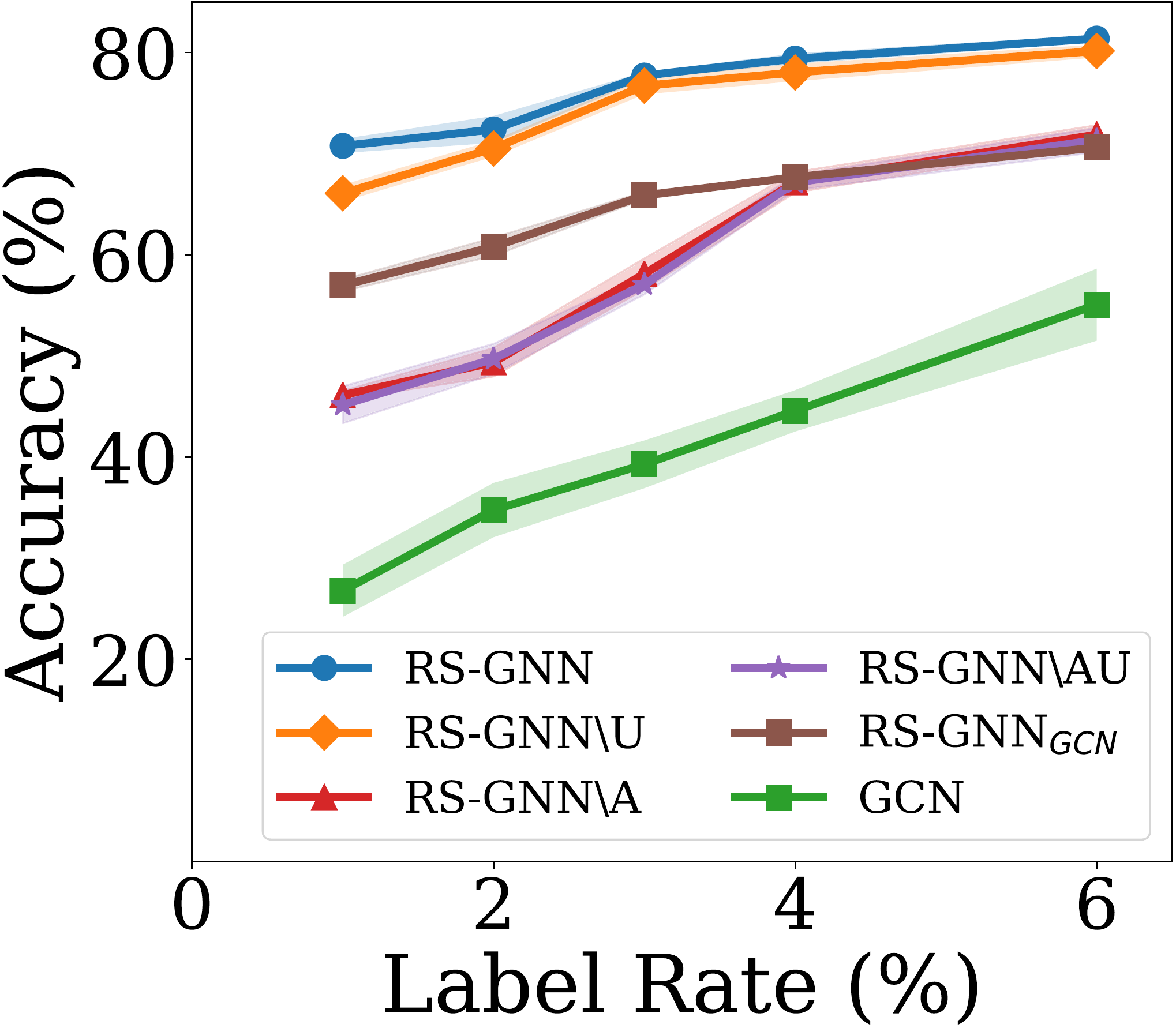} 
    \vskip -0.5em
    \caption{Metattack with 15\% Ptb}
    \label{fig:abla_meta}
\end{subfigure}
\vspace{-1.3em}
\caption{Ablation studies on Cora with different label rates.}
\label{fig:abl}
\vskip -1.8em
\end{figure}
\subsection{Ablation Study}
To answer \textbf{RQ3}, we conduct ablation studies to understand the effects of graph densification, graph purification and label smoothness regularization. In RS-GNN, the link predictor densify the graph to enhance the performance on unlabeled nodes. To demonstrate the effects of adding edges with the link predictor, we remove the process of adding edges and obtain RS-GNN$\backslash$A. 
To testify the effectiveness of the label smoothness regularization based on the generated graph, we eliminate the label smoothness regularization and get RS-GNN$\backslash$U. To show our link predictor can eliminate the effects of noisy edges, we compare a variant named as RS-GNN$\backslash$AU which only use the link predictor to denoise graphs. Graph desification and label smoothness are not applied in RS-GNN$\backslash$AU. 
We also implement a variant named as RS-GNN$_{GCN}$ which uses GCN as link predictor to show that the noisy edges would largely affects the GNNs for link prediction. Hyperparameters selection follows the process in Sec~\ref{sec:implementation}. We only show the results on the Cora graph perturbed with \textit{metattack} and random noise, because similar trends are observed on other datasets. Results are presented in Fig.~\ref{fig:abl}. From this figure, we observe that: 
\begin{itemize}[leftmargin=*]
    \item RS-GNN performs much better than RS-GNN$\backslash$A and RS-GNN$\backslash$U, which shows that densifying graphs and label smoothness with the learned graph can address the label sparsity issue;
    \item With the increase of label rate, the gap between RS-GNN and RS-GNN$\backslash$U will be narrowed. This is consistent with our analysis that higher label rates would involve more unlabeled nodes;
    \item RS-GNN$_{GCN}$ performs much worse than RS-GNN, which indicates adversarial edges would impair GCN and result in a poor link predictor for denoising and densification.
\end{itemize}

\begin{figure}[t]
\centering
\begin{subfigure}{0.49\columnwidth}
    \centering
    \includegraphics[width=0.98\linewidth]{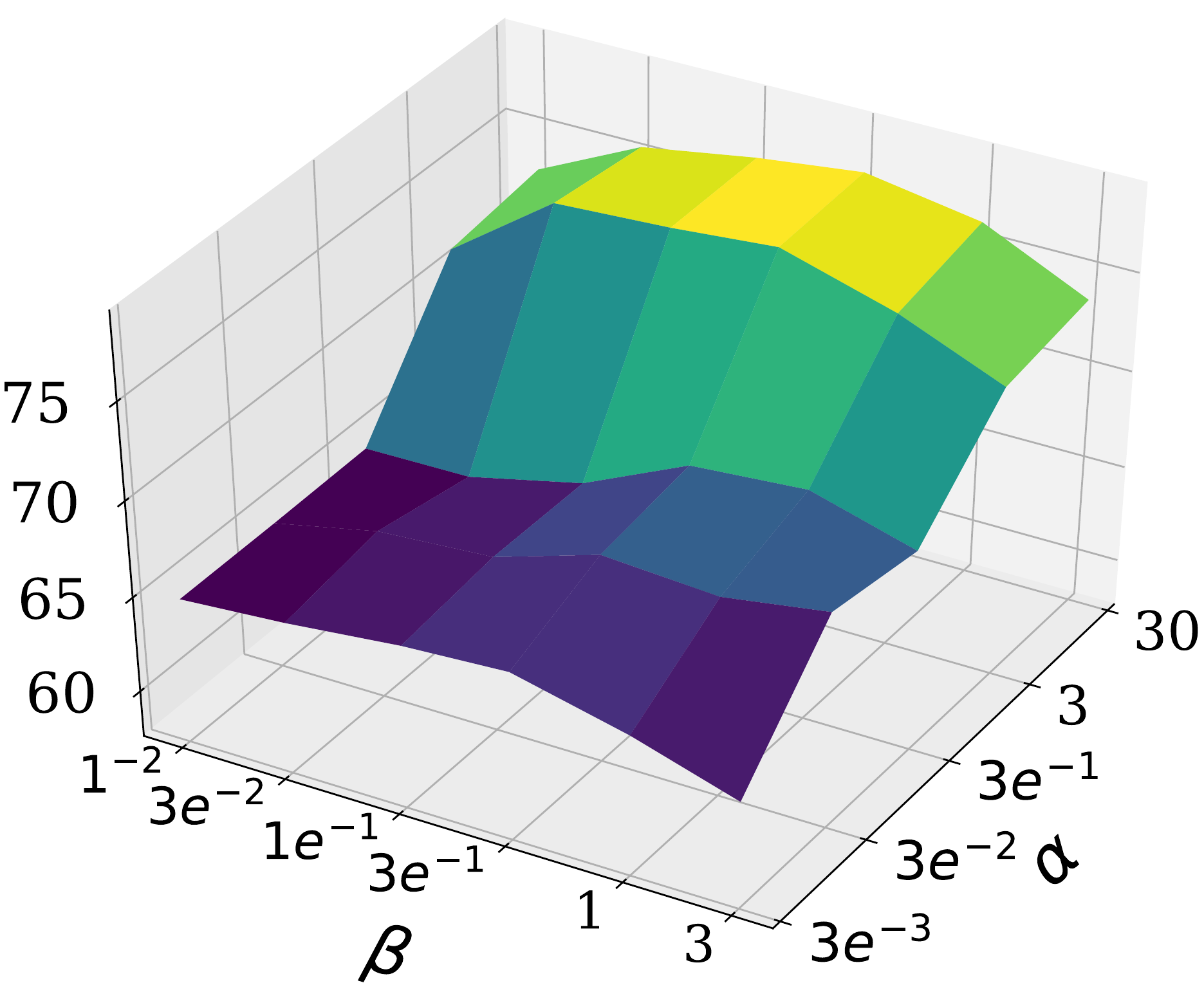}
    \vskip -0.5em
    \caption{Raw Graph}
    \label{fig:para_raw}
\end{subfigure}
\begin{subfigure}{0.49\columnwidth}
    \centering
    \includegraphics[width=0.98\linewidth]{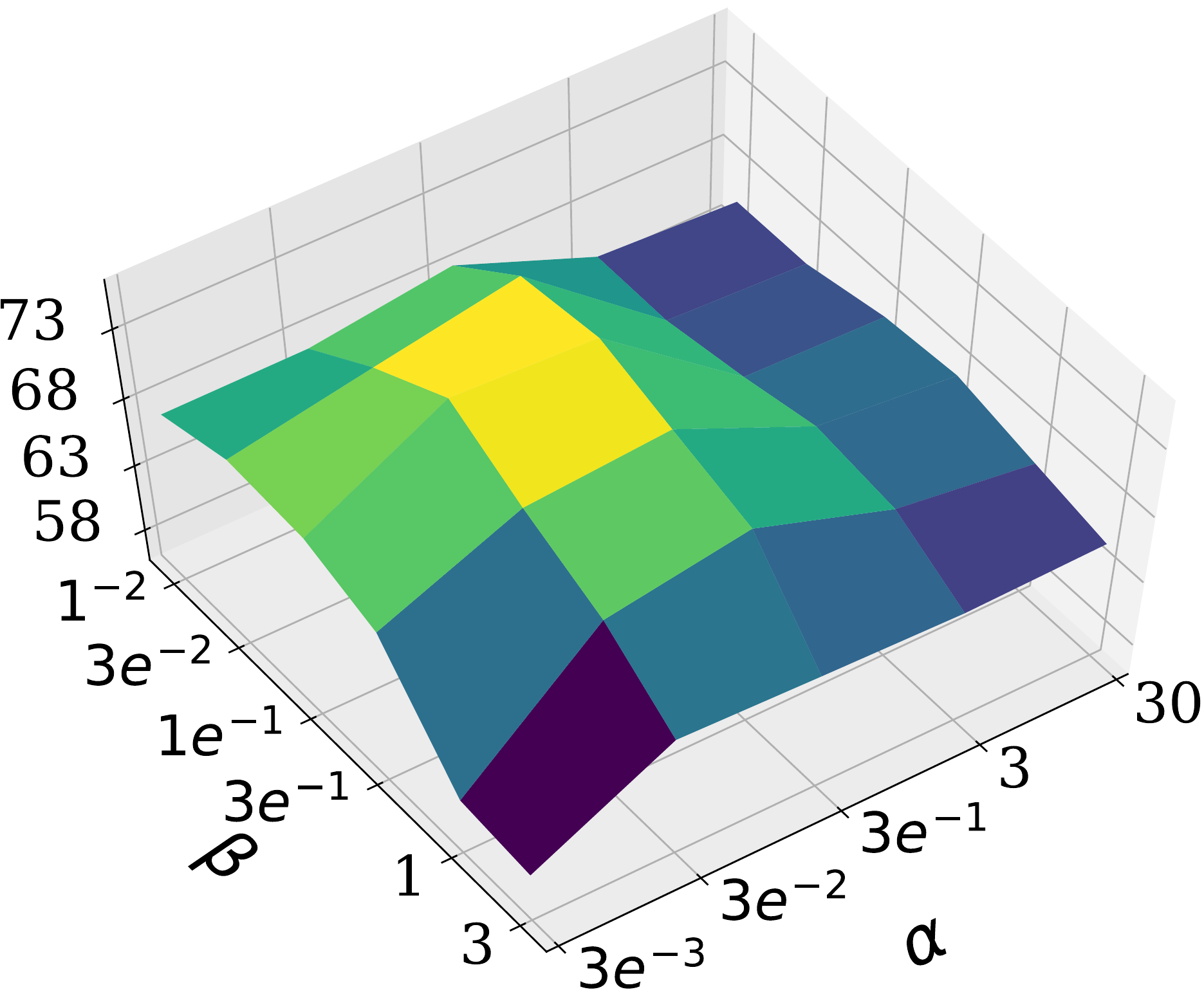}
    \vskip -0.5em
    \caption{Metattack with 15\% Ptb}
    \label{fig:para_meta}
\end{subfigure}
\vspace{-1em}
\caption{Parameter sensitivity analysis on Cora.}
\label{fig:para}
\vskip -1.7em
\end{figure}

\subsection{Parameter Sensitivity Analysis}
\label{Sec:para_analysis}
In this subsection, we explore the sensitivity of the most crucial hyperparameters $\alpha$ and $\beta$ which are in the final objective function of RS-GNN. The analysis about other hyperparameters is presented in the supplementary material. $\alpha$ controls how well the link predictor reconstructs the noisy graph and $\beta$ controls the contribution of label smoothness. To investigate the effects of $\alpha$ and $\beta$, we vary the values of $\alpha$ as $\{0.003, 0.03, 0.3, 3, 30\}$ and $\beta$ as $\{0.01, 0.03, 0.1, 0.3, 1, 3\}$ on Cora. The results are shown in Fig~\ref{fig:para}. In the raw graph, when $\alpha$ is large, the accuracy is stable and high. But if the $\alpha$ is too large in the perturbed graph, the performance would decrease. This difference is due to the noise levels of the raw graph and the perturbed graph. The structural noise in the perturbed graph is severe, faithfully reconstructing the perturbed graph with high $\alpha$ would lead to a poor link predictor. As for the $\beta$, a value between 0.03 to 0.3 generally gives good performance, which eases the parameter selection.

\section{Conclusion and Future Work}
\label{Sec:conclusion}
In this paper, we study a novel problem of learning robust GNNs on noisy graphs with limited labeled nodes. We demonstrate that noisy edges and limited labeled nodes would largely impair the performance of GNNs. A novel RS-GNN is proposed to mitigate these issues. More specially, we adopt the edges in the noisy graph as supervision to obtain a denoised and densified graph to facilitate the message passing for predictions of unlabeled nodes. Moreover, we also utilize the supervision from the generated graph to explicitly involve unlabeled nodes. Extensive experiments on real-world datasets demonstrate the robustness of the proposed framework on noisy graphs with limited labeled nodes. There are several directions requiring further investigation. First, we focus on structural noise in this paper. However, for some applications, such as social networks, users may provide fake attributes for privacy. Thus, we will extend it to graphs with structural noise as well as attribute noise under the setting of limited labeled nodes. Second, the labels may also contain noise which may degrade the performance of GNNs due to the message passing. Therefore, we will also explore methods that handle noisy graphs with limited and noisy labels. 

\section{Acknowledgement}
This material is based upon work supported by, or in part by, the National Science Foundation (NSF) under grant \#IIS1955851, and Army Research Office (ARO) under grant \#W911NF-21-1-0198. The findings and conclusions in this paper do not necessarily reflect the view of the funding agency.

\bibliographystyle{ACM-Reference-Format}
\bibliography{ref}

\newpage
\appendix
\begin{table}[t]
    \small
    \caption{Statistics of datasets.}
    \vskip-1em
    \centering
    \begin{tabularx}{0.98\linewidth}{p{0.15\linewidth}CCCC}
    \toprule
         & Cora & Cora-ML & Citeseer & Pubmed   \\
    \midrule
    \#nodes & 2,485 & 2,810 & 2,110 & 19,717\\
    \#edges & 5,069 & 7,981 & 3,668 & 44,338\\
    \#features & 1,433 & 2,879 & 3,703 & 500\\
    \#classes & 7 & 7 & 6 & 3\\
    \bottomrule
    \end{tabularx}
    \label{tab:dataset}
\end{table}
\section{A Training Algorithm of RS-GNN}
\label{sec:app_alg}
\begin{algorithm}[t!] 
\caption{ Training Algorithm of RS-GNN.} 
\label{alg:Framwork} 
\begin{algorithmic}[1]
\REQUIRE
$\mathcal{G}=(\mathcal{V},\mathcal{E}, \mathbf{X})$, $\mathcal{Y}$, $K$, $Q$ $T_l$, $T_h$, $\sigma$, $\alpha$ and $\beta$.
\ENSURE $f_{\mathcal{G}}$ and $f_E$
\STATE Randomly initialize the parameters of $f_{\mathcal{G}}$ and $f_E$.
\REPEAT 
\STATE Get the denoised and densified graph  $\mathbf{S}$ with $f_E$ by Eq.(4).
\STATE Input the learned graph $\mathbf{S}$ and node attributes $\mathbf{X}$ to GCN classifier $f_{\mathcal{G}}$ to get robust predictions.
\STATE Jointly optimize the GCN classifier parameters $\theta_{\mathcal{G}}$ and the link predictor parameters $\theta_E$ by Eq.(7). 

\UNTIL convergence
\RETURN $f_{\mathcal{G}}$ and $f_E$
\end{algorithmic}
\end{algorithm}
The training algorithm of RS-GNN is presented in Algorithm \ref{alg:Framwork}. In line 1, link predictor $f_E$ and GCN classifier $f_{\mathcal{G}}$ are randomly initialized. In line 2, we generate the graph with $f_E$. 
Then the link predictor and GCN classifier are jointly trained in an end-to-end manner by Eq. (7) in line 3. 
Adam optimizer with learning rate set as 0.001 is applied to update all the parameters.

\begin{table}[h]
    \small
    \centering
    \caption{Number of involved unlabeled nodes}
    \vskip -1em
    \begin{tabularx}{0.95\linewidth}{p{0.24\linewidth}XXXX}
    \toprule
        Dataset &  Cora & CoraML & Citeseer & Pubmed \\
    \midrule
        Raw Graph&  212 & 447 & 168 & 12,430 \\
        Generated Graph & 1,383 & 2,161 & 955 & 18,555\\
    \bottomrule
    \end{tabularx}
    \label{tab:neighbs}
\end{table}
\section{More details of the Learned Graph}

\label{sec:app_graph}
Since RS-GNN aims to densify the graphs to benefit predictions in sparsely labeled graphs, we compare the number of involved unlabeled nodes in raw and generated graphs. More specially, in a two layer GNN, the neighbors of labeled nodes within two hops will participate in the training process. The generated graphs are attained by training RS-GNN on graphs perturbed by random noise. We binarize weighted edges by setting 0.5 as the threshold. The comparisons are given in Table \ref{tab:neighbs}. We can find that more unlabeled nodes are involved in the training with the generated graphs, which implies that RS-GNN could promote predictions of unlabeled nodes by densifying graphs.
\section{The Impacts of Hyperparmeters}
\label{sec:app_hyperparameter}

\begin{table}[t]
    \small
    \centering
    \caption{The impacts of hyperparameter $K$.}
    \vskip -1em
    \begin{tabularx}{0.88\linewidth}{XCCCC}
    \toprule
    $K$  &  50 & 100 & 200 & 400    \\
    \midrule
    Cora &  66.4 $\pm 1.8$ & \textbf{70.8} $\pm \textbf{0.7}$ & 69.5 $\pm 2.9$ & 68.2 $\pm 3.3$\\
    Cora-ML & 44.8 $\pm 1.2$ & 53.8 $\pm 2.7$ & \textbf{73.2} $\pm \textbf{1.2}$ & 69.0 $\pm 5.0$\\
    Citeseer & 63.3 $\pm 2.0$ & 66.0 $\pm 1.4$ & \textbf{68.0} $\pm \textbf{0.4}$ & 67.8 $\pm 1.4$\\
    \bottomrule
    
    \end{tabularx}

    \label{tab:para_K}
\end{table}

\begin{table}[t]
    \small
    \centering
    \caption{The impacts of hyperparameter $T_h$.}
    \vskip -1em
    \begin{tabularx}{0.88\linewidth}{XCCCC}
    \toprule
    $T_h$  &  0.6 & 0.7 & 0.8 & 0.9    \\
    \midrule
    Cora &  68.3 $\pm 0.7$ & 68.9 $\pm 1.5$ & \textbf{70.8} $\pm \textbf{0.7}$ & 69.8 $\pm 2.1$\\
    Cora-ML & 64.8 $\pm 4.1$ & 68.2 $\pm 3.6$ & \textbf{73.2} $\pm \textbf{1.2}$ & 69.2 $\pm 4.8$ \\
    Citeseer & 66.6 $\pm 1.7$ & 67.5 $\pm 2.1$ & \textbf{68.0} $\pm \textbf{0.4}$ & 67.8 $\pm 2.2$\\
    \bottomrule
    
    \end{tabularx}

    \label{tab:para_t}
\end{table}

\begin{table}[t]
    \small
    \centering
    \caption{The impacts of hyperparameter $T_l$.}
    \vskip -1em
    \begin{tabularx}{0.88\linewidth}{XCCCC}
    \toprule
    $T_l$  &  0.0 & 0.05 & 0.1 & 0.2   \\
    \midrule
    Cora &  65.5 $\pm 2.8$ & 68.5 $\pm 3.3$ & \textbf{70.8} $\pm \textbf{0.7}$ & 70.3 $\pm 1.4$\\
    Cora-ML & 65.9 $\pm 2.6$ & 72.5 $\pm 1.3$ & \textbf{73.2} $\pm \textbf{1.2}$ & 69.6 $\pm 3.9$ \\
    Citeseer & 65.8 $\pm 0.6$ & 66.8 $\pm 0.8$ & \textbf{68.0} $\pm \textbf{0.4}$ & 66.6 $\pm 1.3$\\
    \bottomrule
    
    \end{tabularx}

    \label{tab:para_l}
\end{table}

\begin{table}[t]
    \small
    \centering
    \caption{The impacts of hyperparameter $\sigma$.}
    \vskip -1em
    \begin{tabularx}{0.88\linewidth}{XCCCC}
    \toprule
    $\sigma$  &  30 & 100 & 300 & 1000    \\
    \midrule
    Cora &  70.2 $\pm 1.2$ & \textbf{70.8} $\pm \textbf{0.7}$ & 70.1 $\pm 1.1$ & 68.9 $\pm 2.8$\\
    Cora-ML & 72.7 $\pm 1.0$ & \textbf{73.2} $\pm \textbf{1.2}$ & 72.5 $\pm 0.8$ & 72.4 $\pm 0.5$ \\
    Citeseer & 66.1 $\pm 1.3$ & \textbf{68.0} $\pm \textbf{0.4}$ & 67.3 $\pm 0.9$ & 66.5 $\pm 1.0$\\
    \bottomrule
    
    \end{tabularx}

    \label{tab:para_sigma}
\end{table}
\textbf{Impacts of $K$.}
When we add edges with the link predictor, for each node, we select $K$ nodes with the largest cosine similarity as candidate node set to predict the links to reduce the computational cost. To investigate how the selection of $K$ would influence the training, we vary $K$ as \{50, 100, 200, 400\} and report the average accuracy of 5 runs on Cora, Cora-ML and Citeseer that are perturbed by \textit{metattack} in Table~\ref{tab:para_K}. The perturbation rate is set as 0.15. The label rate is set as 0.01 which is the same as that of main paper. We can observe that with the increase of $K$, the performance would firstly increase a lot then slightly decrease. Because when $K$ is small, there are not adequate candidate nodes to predict links for each node. In this situation, the learned graph will be still sparse, which leads to poor performance on the noisy graphs with sparse labels. When $K$ is very large, for a node $v$, nodes that dissimilar with $v$ in raw features space would also be added into the candidate set. As a result, the performance slightly decrease.

\textbf{Impacts of $T_h$.}
When we apply the label smoothness regularization based on the generated graph, we will smooth the predictions of nodes linked by predicted links whose weights are larger than $T_h$.
To investigate how the setting of $T_h$ affects the label smoothness regularization, we vary $T_h$ as  $\{0.6,0.7,0.8,0.9\}$. 
We conduct experiments on the graphs perturbed by \textit{metattack}. The perturbation rate is set as 0.15. The label rate is set as 0.01. Other parameters follows the same settings in the main paper. Average results of 5 runs are reported in Table~\ref{tab:para_t}.  It shows that $T_h$ should be set as an appropriate value such as 0.8 to benefit the predictions with label smoothness.

\textbf{Impacts of $T_l$.} When we deniose and desify the graph, a $T_l$ is applied to the results of link predictor to determine whether we should keep/add the links. We vary the value of $T_l$ as $\{0.0, 0.05, 0.1, 0.2\}$ to investigate the influence of $T_l$. Experiments are conducted on the graphs perturbed by \textit{metattack} with the perturbation rate set as 0.15. The average results of 5 runs are reported in Table~\ref{tab:para_l}.  As we can see, with the increase of $T_l$, the performance will firstly increase and then decrease. Because when $T_l$ is very small, a lot of down-weighted noisy edges are not removed, which degrades the performance of RS-GNN. If $T_l$ is too large, the size of assigned links will be limited and some normal edges are likely to be deleted. Thus, the performance will drop when $T_l$ is too large.

\textbf{Impacts of $\sigma$.} In Eq.(3) of our main paper, a parameter $\sigma$ is used to control the variance of the weights of positive samples and negative samples when we train the link predictor with the loss of reconstructing the noisy graph. We vary the value of $\sigma$ as $\{30, 100, 300, 1000\}$ and fix other hyperparameters. Similarly, experiments are conducted of the Cora, Cora-ML, and citeseer graphs perturbed by \textit{metattack} with the perturbation rate set as 0.15. The results are presented in Table~\ref{tab:para_sigma}. From this table we could observe that when the $\sigma$ is set too large, the performance will decrease. When $\sigma$ is very large, the weights of all the negative samples and positive samples will be similar, which results a poor link predictor affected by noisy edges. This demonstrates the effectiveness of reweighting the samples based on raw feature similarity. However, if the $\sigma$ is too small, the variance of sample weights would be too large, which negatively affects the learning of link predictor.

\end{document}